\documentclass[10pt,twocolumn,letterpaper]{article}

\usepackage[final]{cvpr}      

\usepackage{graphicx}
\usepackage{amsmath}
\usepackage{amssymb}
\usepackage{bm}
\usepackage{booktabs}

\usepackage[pagebackref,breaklinks,colorlinks]{hyperref}

\usepackage[capitalize]{cleveref}
\crefname{section}{Sec.}{Secs.}
\Crefname{section}{Section}{Sections}
\Crefname{table}{Table}{Tables}
\crefname{table}{Tab.}{Tabs.}

\def\cvprPaperID{*****} 
\def\confName{CVPR}
\def\confYear{2022}

\usepackage{algorithm}
\usepackage{algorithmicx}
\usepackage{algpseudocode}

\usepackage{ifthen}
\usepackage{xifthen}
\def\*#1{\mathbf{#1}} \def\+#1{\mathcal{#1}} \def\-#1{\mathrm{#1}}\def\^#1{\mathbb{#1}}\def\!#1{\mathtt{#1}}
\newcommand{\set}[1]{\left\{#1\right\}}

\renewcommand{\mid}{\;\middle\vert\;}

\renewcommand{\Pr}[2][]{ \ifthenelse{\isempty{#1}}
  {\mathbf{Pr}\left[#2\right]} {\mathbf{Pr}_{#1}\left[#2\right]} }
\newcommand{\E}[2][]{ \ifthenelse{\isempty{#1}}
  {\mathbb{E}\left[#2\right]}
  {\mathbb{E}_{#1}\left[#2\right]} }
\newcommand{\Var}[2][]{ \ifthenelse{\isempty{#1}}
  {\mathbf{Var}\left[#2\right]}
  {\mathbf{Var}_{#1}\left[#2\right]} }
\newcommand{\Ent}[2][]{ \ifthenelse{\isempty{#1}}
  {\mathbf{Ent}\left[#2\right]}
  {\mathbf{Ent}_{#1}\left[#2\right]} }

\newcommand{\dist}[2][]{ \ifthenelse{\isempty{#1}}
  {\mathrm{dist}\left[#2\right]}
  {\mathrm{dist}_{#1}\left[#2\right]} }

\newcommand{\norm}[2][]{ \ifthenelse{\isempty{#1}}
  {\left\|#2\right\|}
  {\left\|#2\right\|_{#1}} }

\begin{document}

\title{Facke: a Survey on Generative Models for Face Swapping}

\author{Wei Jiang\footnotemark[1], Wentao Dong\footnotemark[1]\\
Shanghai Jiaotong University, Shanghai, China\\
{\tt\small \{ailon\_jw, a\_Dong\}@sjtu.edu.cn}\\
\url{https://github.com/Ailon-Island/Facke}
}

\maketitle

\renewcommand{\thefootnote}{\fnsymbol{footnote}}
\footnotetext[1]{Equal contribution.}
\renewcommand{\thefootnote}{\arabic{footnote}}

\section{Introduction}
    Generation of realistic objects (images, text, sound, etc.) is one of the keys to an authentic representation of the world. Though our observation of the world is always limited and expensive, with generative models we can supplement or even refine a reliable observation for later utilization in fields like data science, computer perception, reinforce learning, and virtual reality. To make sure the validity of the supplement, we tend to generate objects from the identical distribution of the original observation. In this sense, most existing methods approximate the distribution and synthesize novel objects by sampling. To facilitate sampling, dense sample spaces are preferred, and the latent samples are later projected onto the space of explicit representation. Neural generators, \eg, Variational Auto-encoder (VAE, \cite{vae}) , Generative Adversarial Network (GAN, \cite{gan}), and Denoising Diffusion Probabilistic Model (DDPM, \cite{ddpm}), have been proposed to fulfill this purpose. These models are demanding in computation, but could provide far higher fidelity and diversity.
    
    As is inferred by the name Facke (fake face), in this project, we focus on the very task of face swapping, i.e., replacing the face in the target image with that in the source one. A widely adopted idea is to utilize conditioned generators (especially neural generators) of images. In this pipeline, personal identity information from the source image is injected during the resampling (from image space to sample space, and back to image space) of the target image. Face swapping with neural generators has been prevailing since Deepfakes published the first prototype of Faceswap \cite{deepfakes} in 2017. Most neural generative models have been tried in this task, and the current mainstream models are Conditional Variational Auto-encoder (CVAE, \cite{cvae}), Conditional Generative Adversarial Network (CGAN, \cite{cgan}), and CVAE-GAN\cite{cvaegan}, which is the combination of the former two. Existing finely trained models have already managed to produce fake faces indistinguishable to the naked eye. And emerging methods, especially conditioning methods based on DDPM (\eg, ILVR\cite{ilvr}, LDM\cite{ldm}, etc.), are promising to generate images with even higher fidelity.
    
    In specific, we investigate into the three typical methods of CVAE, CGAN, and ILVR. CVAEs tend to spawn images with vague edges, and are prone to ill-shot faces (\eg, profiles, and shaded faces). With adversarial self-supervision component, CGANs allow various penalties, and utilize identity information more properly. However, their instability makes it a hard time for longer training schedules. Feature matching has been adapted to stabilize the model and avoid mode collapse. To reserve diversity of result during feature matching, Bao \etal proposed CVAE-GAN in \cite{cvaegan}, is a hybrid of CVAE and GAN. Another method of ILVR is also explored. It is relatively poor at identity retrieving, but excels in image quality. And we have to mention the significantly slower speed of this method.

\section{Method}
\subsection{GAN}
Generative Adversarial Network \cite{gan} is a clever framework of self-supervised neural generators. Instead of providing some concrete structures of a generator, it gives a philosophy to train generators. A typical GAN model consists of a generator and a discriminator (or classifier). The discriminator is trained jointly with the generator so as to serve as a learning supervisor. Balanced learning speed ensured, both the generator and the discriminator (or classifier) will make steady improvement along the gambling.
\subsubsection{Vanilla GAN}
Let us consider a generator $G(\boldsymbol{z}; \theta_g)$ which projects noise $\boldsymbol{z}$ in latent space into the object space, \ie, the face image space. And we define a discriminator $D(\boldsymbol{x};\theta_d)$ that outputs the probability that $\boldsymbol{x}$ is real . (We say an image $\boldsymbol{x}$ is real if it is sampled from data rather than generated by $G$.) Simultaneously, we train $D$ to maximize the probability of correct discrimination between real and fake images, and train $G$ to minimize the probability that $D$ makes wrong judgement on whether $G(\boldsymbol{x})$ is fake. So this is a minmax game with value function
\begin{equation}
    \begin{aligned}
        \min_G \max_D V(D,G) &= \E[\boldsymbol{x}\sim p_{\mathrm{data}}(\boldsymbol{x})]{\log D(\boldsymbol{x})} \\
        &+ \E[\boldsymbol{z} \sim p_{\boldsymbol{z}}(\boldsymbol{z})]{\log {(1-D(G(\boldsymbol{z})))}}
    \end{aligned}
\end{equation}

As is shown by \cite{gan}, the training criterion allows recovery of the data generating distribution given enough capacity.
\subsubsection{Image to Image}
In our face swapping task, we would prefer an end-to-end model, \ie, the model should take in at least one image and output an image (swapped face). The input $\boldsymbol{z}$ in the latent space certainly cannot meet this requirement, so we turn to an Auto-encoder as the generator so as to project a face into a latent $\boldsymbol{z}$ and then back to an image. At the bottleneck of the model, \ie, in latent space of data, we tend to get a highly compressed and informative representation. 
\subsubsection{CGAN and SimSwap}
\begin{figure*}
  \centering
    \includegraphics[width=\linewidth]{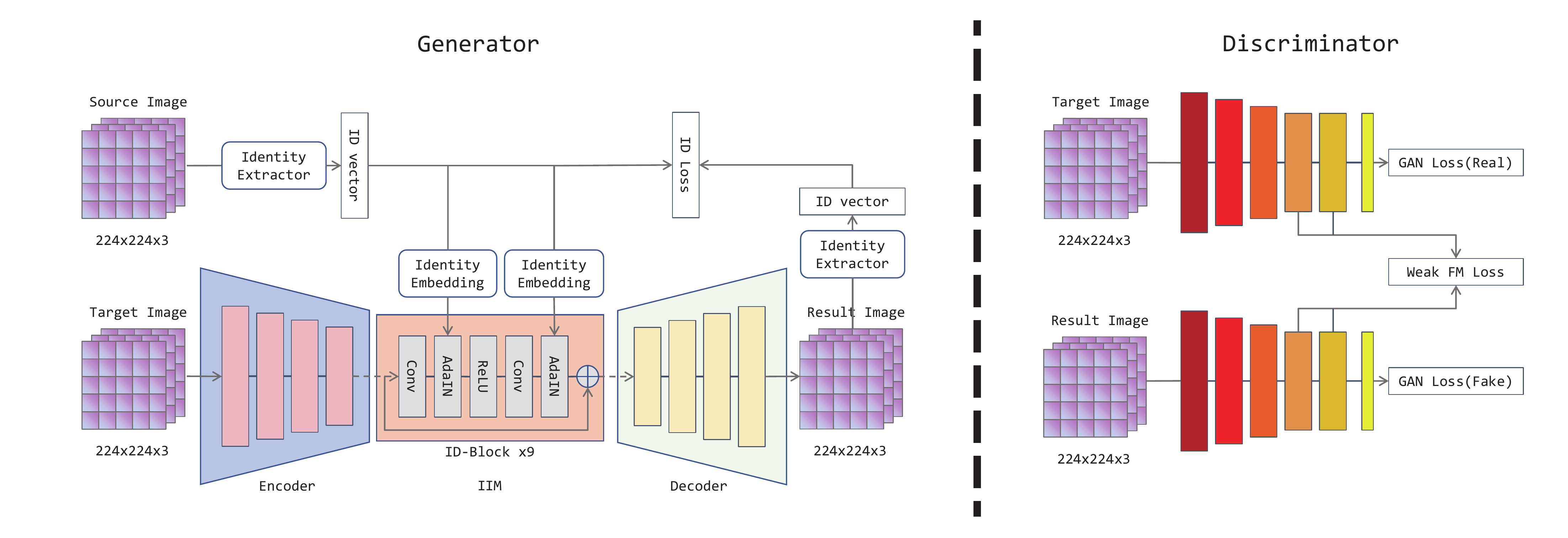}
    \caption{Architecture of Original SimSwap.}
    \label{fig:simswap}
\end{figure*}

Now we have scratched a framework of how to generate a face from the target face. Then we need to somehow inject the identity of a source face during generation. A straightforward solution is to use the conditioned version of GAN where we condition the model with source identity, which is what SimSwap \cite{simswap} and other CGAN-based works do. We strictly follow SimSwap to build our network. Instead of taking source face directly as a condition, SimSwap extracts an ID vector from it by pretrained ID extractor \cite{arcface}. And the conditioning, \ie, ID injection, is done by a series of ResNet\cite{resnet}-like blocks called ID Blocks, in which the latent vector gets reprojected and injected ID with AdaIN \cite{adain} repeatedly. Such a design enables critical feature detection and dedicated modification. The architecture of SimSwap network is shown in \cref{fig:simswap}. Note that SimSwap actually has two collaborating multi-scale discriminators $D_1$ and $D_2$ to better performances under large postures.

In SimSwap, the overall loss is defined in the following manner:
\begin{equation}
\begin{aligned}
    \+L =& \lambda_{Id}\+L_{Id} + \lambda_{Recon}\+L_{Recon} + \+L_{Adv}\\ 
    +& \lambda_{GP}\+L_{GP} + \lambda_{wFM}\+L_{wFM}
\end{aligned}
\end{equation}
where $\lambda_{Id} = 30, \lambda_{Recon} = 10, \lambda_{GP} = 10^{-5}, \lambda_{wFM} = 10 $. And the losses are Identity Loss, Reconstruction Loss, Hinge version \cite{hingeloss} of Adversarial Loss, Gradient Penalty Loss \cite{wgan, wgan-gp}, and Weak Feature Matching Loss, respectively. The definition of these losses are 
\begin{align}
    &\+L_{Id} = 1 - \frac{\left\langle v_R, v_S\right\rangle}{\norm[2]{v_R} \norm[2]{v_S}}\\
    &\+L_{Recon} = \norm[1]{I_R - I_T}\\
    &
    \begin{aligned}
    \+L_{Adv\_D}=&\E{\max\set{0, 1-D(\boldsymbol{x})}}\\
    +& \E{\max \set{0, 1+D(G(\boldsymbol{z}))}}
    \end{aligned}\\
    &\+L_{Adv\_G} = -\E{D(G(\boldsymbol{z}))}\\
    &\+L_{GP}=\left( \norm[2]{\nabla_{\hat{\boldsymbol{x}}} D(\hat{\boldsymbol{x}})}-1\right)^2\\
    &\+L_{wFM}=\sum_{i=m}^M \frac{1}{N_i} \norm[1]{D^{(i)} (I_R) - D^{(i)}(I_T)  } \label{loss:wFM}
\end{align}
where $\hat{\boldsymbol{x}} = \epsilon \boldsymbol{x} - (1-\epsilon) \Tilde{\boldsymbol{x}}, \epsilon \sim U[0,1]$. Different from original Featrue Matching proposed in pix2pixHD \cite{pix2pix}, with no ground truth for face swapping, SimSwap only includes the last few layers in the calculation. In \cref{loss:wFM}, $D^{(i)}$ denotes the $i$-th layer of discriminator $D$, $N_i$ denotes the number of elements in $D^{(i)}$, and $m$ is the layer to start calculating the Feature Matching Loss. By deprecating the first few layers, Weak Feature Matching makes the network learn to preserve attributes from target faces without introducing constraints on detailed textures (which may lead to identity alignment).

Note that we do not optimize $\+L_{Recon}$ for "inter-ID batches".

\subsubsection{Identity Grouping}
In the training phase, we follow SimSwap to train one batch for image pairs with the same identity (we call it an intra-ID batch) and another batch for those with different identities (an inter-ID batch, as a counterpart). We propose to use a data sampling schedule of Identity Grouping to boost the network's ability of choosing precise Identity information from ID vectors. In original SimSwap \cite{simswap}, intra-ID batches takes exactly the same source face and target face. Some datasets (\eg, VGGFace2-HQ \cite{VGGFace2}) may have a collection of images for one identical person. So, in every intra-ID batch, we first sample $b$ (batch size) peoples (classes) uniformly, and then uniformly pick two images for each selected person as source face and target face. That means, the model is self-swapping between faces of the same identity but with different attributes. (We make great efforts to get this compatible with multiple workers.)

This trick seems hopefully to help model tell attributes from identity in the ID vectors. Furthermore, we have found that the ID extractor \cite{arcface} is not strong enough to give sufficiently close ID vectors for images of the same identity. With Identity Grouping, the model tends to output faces with ID vectors within the distribution of ID vectors of the very source identity. Note that the optimum is the spherical mean of the distribution, \ie, our model is optimized to find a corresponding stable identity center from an ID vector. Since the ID vectors of an identity situates around the pure identity and vibrates with remaining attributes, we can claim that our model could further purify the identity information from an ID vector so as to inject identity more exactly.

\subsection{CVAE-GAN}
\subsubsection{VAE}
    Varational Auto-Encoder \cite{vae} is a directed graphical model with certain types of latent variables. We force the prior of the latent variables $\boldsymbol{z}$ is standard Gaussian distribution to ensure the latent space is well distributed. Because of the intractability of the trure posterior $p_\theta(\boldsymbol{z}| \boldsymbol{x})$, we introduce a recognition model $q_\phi (\boldsymbol{z} | \boldsymbol{x})$ as a probabilistic \emph{encoder}. In the process of generation of new $\boldsymbol{x}$ from latent variable $\boldsymbol z$, we want that $\boldsymbol{x}$ is generated from a conditional distribution $p_\theta (\boldsymbol{x}| \boldsymbol{z})$, which is the probabilistic \emph{decoder}.
    
    For any $\boldsymbol z$ sent into the decoder, we will not directly use the output of the encoder. Our encoder will output an distribution $\mathcal{N}(\boldsymbol{\mu}, \boldsymbol{\sigma^2})$, and we will apply the \emph{reparameterization} trick to sample $\boldsymbol z$ from the distribution we got from the encoder.
    
    For the loss function, it is formed by two parts, Reconstruction loss and posterior regulation loss. In addition, we use a quite small weight for the regulation term to ensure the latent variables are characteristic in the first place.
    \begin{equation}
        \+L = \mathbb{E}_{\boldsymbol z\sim q_\phi(\boldsymbol z|\boldsymbol x)}[\log(p_\theta(\boldsymbol x | \boldsymbol z))] + D_{\text{KL}}(q_\phi (\boldsymbol z |\boldsymbol x) \| p(\boldsymbol z))
    \end{equation}
    In the train process, we simply use the MSE loss between the original image and the generated image to represent the reconstruction loss.
\subsubsection{CVAE}
    Conditional Varaitonal Auto-Encoder can be regarded as a conditional version of VAE, which means that any probability we may use should baseed on the condition from the source image. We have the encoder $q_\phi(\boldsymbol y | \boldsymbol x , \boldsymbol y)$, the decoder $p_{\theta} (\boldsymbol x | \boldsymbol z, \boldsymbol y)$. 
    
    The difference in programming is quite trivial. We only need to put the source image $\boldsymbol y$ together with the target image $\boldsymbol x$ into the encoder as inputs, and also put the $\boldsymbol y$ together with latent variables $\boldsymbol z$ into the decoder as the input.
\begin{figure*}
  \centering
  \begin{subfigure}{0.5\linewidth}
    \includegraphics[width=\linewidth]{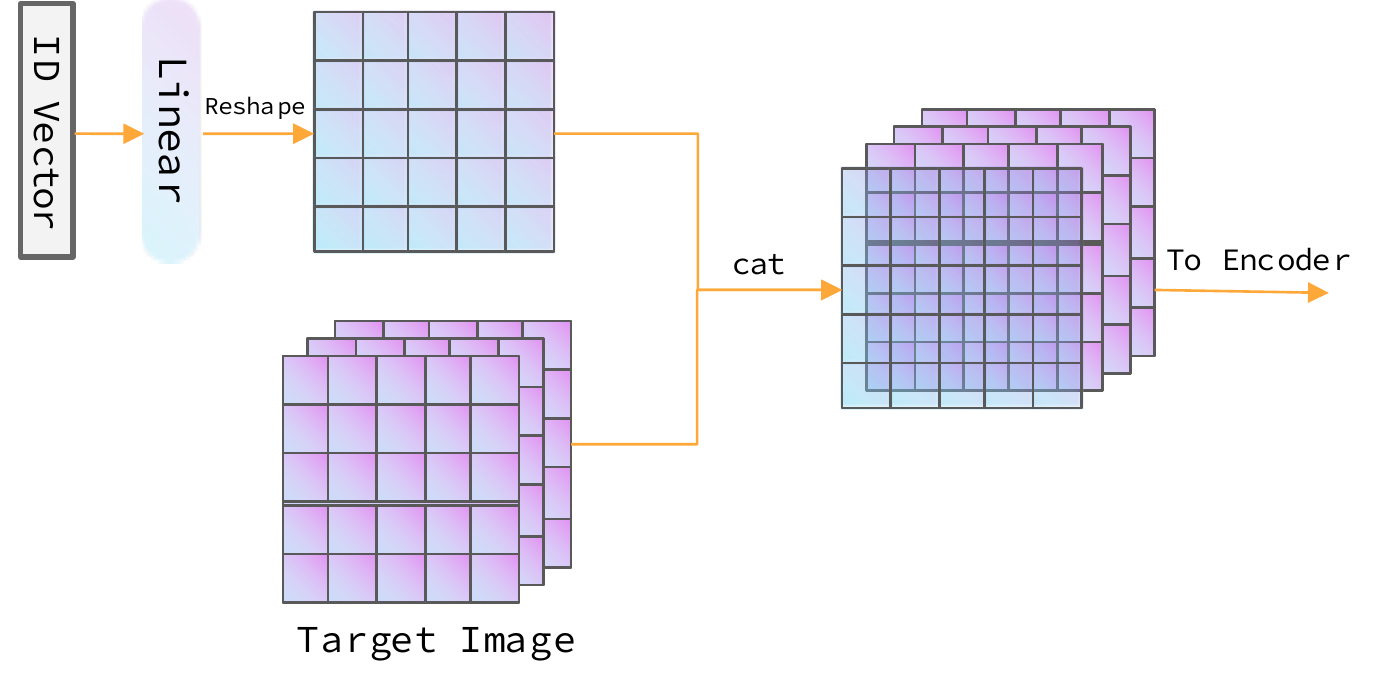}
    \label{fig:CVAE1}
    \end{subfigure}
    \hfill
    \begin{subfigure}{0.4\linewidth}
    \includegraphics[width=\linewidth]{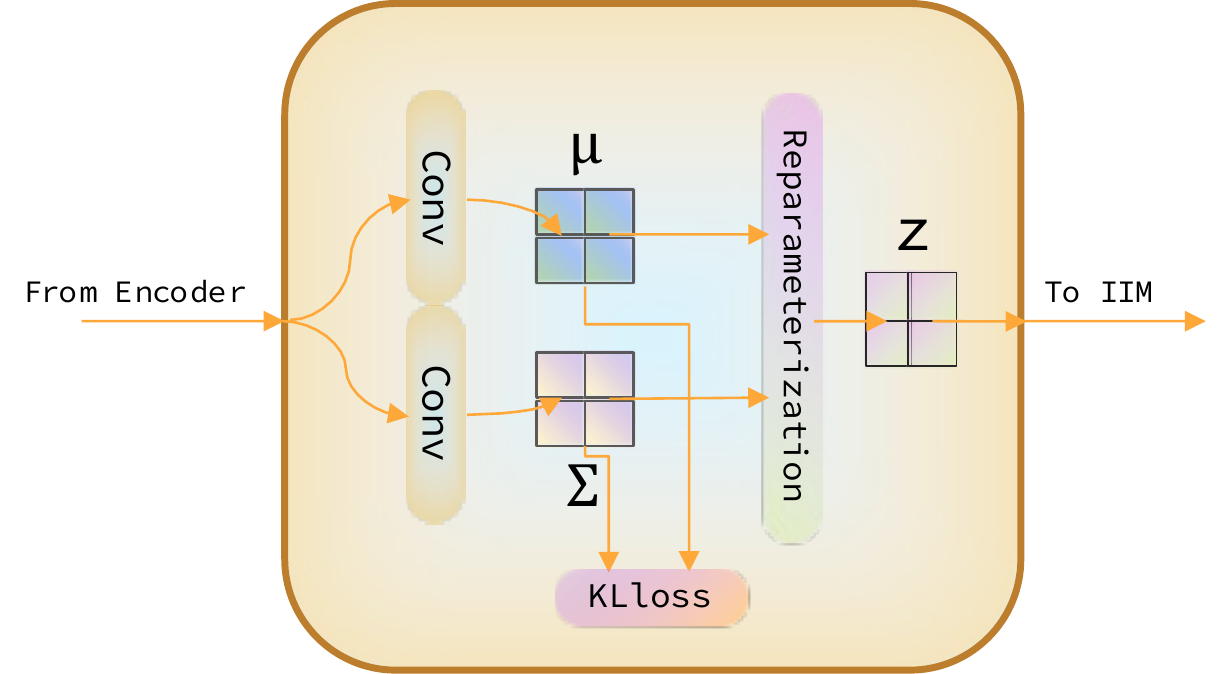}
    \label{fig:CVAE2}
    \end{subfigure}
    \caption{Some Extra Components of CVAE Network Compared with the Original Generator}
    \label{fig:CVAE}
\end{figure*}

\subsubsection{2D Latent}
    To deal with the problem that vallina CVAE only generate blurry image, we turned the linear layers before reparameterization into convolutional ones, which is aiming to restore some spatial information into the latent variables. As a result, we obtain a latent 2D image space. In this way, we are allowed to use the same method of feature injection as SimSwap, \ie, using ID Blocks to AdaIN the ID vector into the reparameterized latent variables $\boldsymbol{z}$.

\subsubsection{Identity Loss}
    From some simple experiments, we found that the source image didn't play its role in the process of face swapping. So we introduce the Identity Loss to make it work. The training schedule of CVAE with Identity Loss is similar with CGAN. For odd iterations, we train the net with the same target image and source image with a higher weight of reconstruction loss. And for even iterations, we train the net with different target image and source image with a lower weight of reconstruction loss. 
    
    With all architecture details mentioned above, our CVAE-GAN's architecture looks quite similar to SimSwap. There are only 2 extra components compared to the original framework of SimSwap, as is shown in \cref{fig:CVAE}
\begin{figure*}
  \centering
  \begin{subfigure}{0.3\linewidth}
    \includegraphics[width=\linewidth]{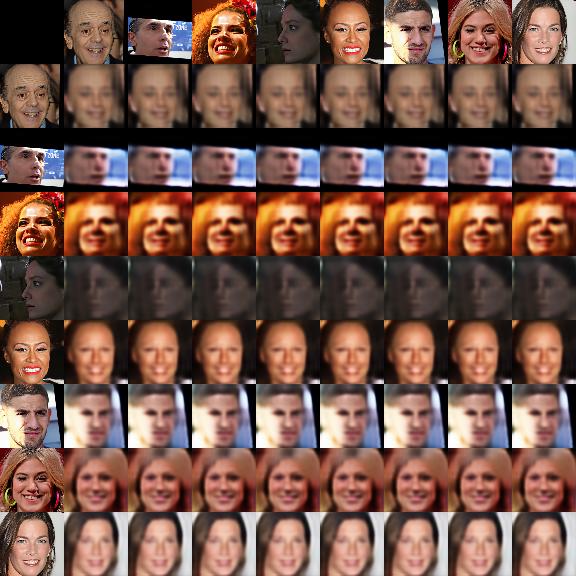}
    \caption{Blurry Image Generated by Original CVAE}
    \label{fig:CVAE_BLUR}
    \end{subfigure}
    \hfill
    \begin{subfigure}{0.3\linewidth}
    \includegraphics[width=\linewidth]{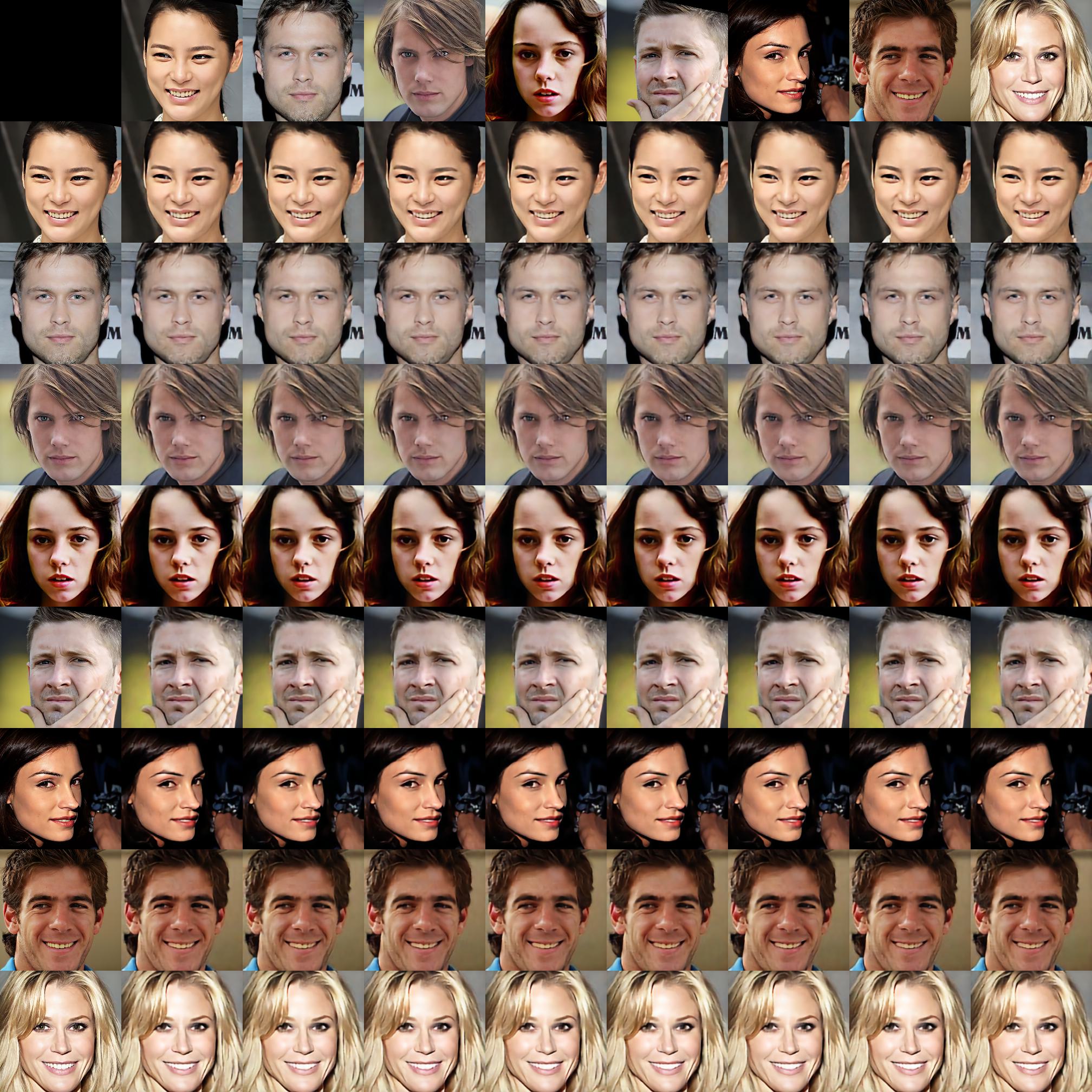}
    \caption{Clear Image with New Conv-Framework but No Face Swapping}
    \label{fig:CVAE_CLEAR}
    \end{subfigure}
    \hfill
    \begin{subfigure}{0.3\linewidth}
    \includegraphics[width=\linewidth]{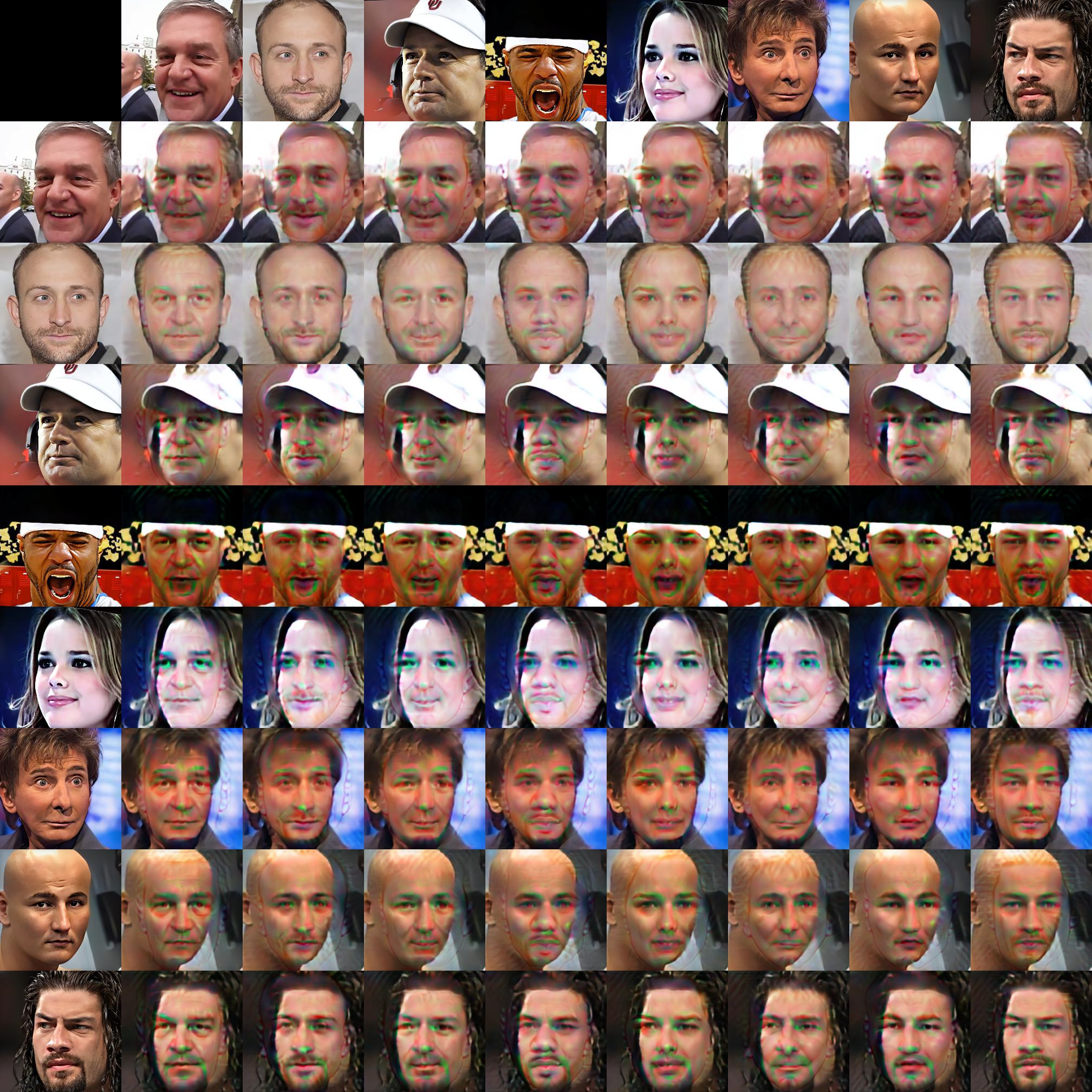}
    \caption{The "Paste" Identity Image Generated by CVAE with IDloss}
    \label{fig:CVAE_Paste}
    \end{subfigure}
    \caption{The Poor Results of non-GAN CVAE Model}
    \label{fig:CVAE_OUT}
\end{figure*}
    
\subsubsection{CVAE-GAN}
    CVAE models are generating "pasted" faces with vague edges and normalized shape. Without supervision over fidelity, the model is fooling the ID extractor to generate ID vectors with lowest Identity Loss through feeding complete, unbiased posed faces regardless of the target posture, shown in \cref{fig:CVAE_OUT}. As a final solution, we follow \cite{cvaegan} to get CVAE and GAN work together, and let a discriminator to ensure the output quality. 
    
    In the paper of CVAE-GAN \cite{cvaegan}, there is an extra part of classifier network C to measure the class probability of the data, which can betterment the category info of the generated image. Due to the limitation of time and GPU resources, we simply replace the generator in CGAN with the CVAE model to get the final model without introducing the classifier. (Otherwise we have to re-scale the images or decrease the batchsize, which means we need 2 times amount of time to train and we may need to reset some of the hyper parameters of the network.) Actually,  we can see the Identity Loss, Reconstruction Loss, and Weak Feature Matching loss as a substitute of a classifier since they help to measure the distance from an output to the objective identity and attributes, which are the potential "classes" in our task.

\subsection{DDPM-ILVR}
Another alternative is DDPM, which is inspired by nonequilibrium thermodynamics. By simulating the thermal diffusion and the reverse counterpart, DDPM allows image generation from noise as well as output change by modifying intermediate results with extra methods. And ILVR is among the latter. Different from methods mentioned above, ILVR is a source-oriented method, which means we blend attributes of the target face to the source one.
\subsubsection{DDPM}

    DDPM\cite{ddpm} is based on Diffusion Probabilistic Model (DPM)\cite{dpm} which is a parameterized Markov chain trained using variational inference to produce samples matching the data after finite time. For each step of the Markov chain, small amounts of Gaussian noise will be added into the the data until the graph sigmal is totally destroyed. Such a process is called \emph{forward process} or \emph{diffusion process}. For this process, we have
    \begin{equation}
        q(\boldsymbol{x}_{1:T}|\boldsymbol{x}_0):= \prod_{t=1}^T q(\boldsymbol{x}_t|\boldsymbol{x}_{t-1})
    \end{equation} and 
    \begin{equation}
        q(\boldsymbol{x}_t | \boldsymbol{x}_{t-1}) := \mathcal{N}(\boldsymbol{x}_t; \sqrt{1-\beta_t} \boldsymbol{x}_{t-1}, \beta_t \boldsymbol{I})
    \end{equation}
    $\beta_1,\dots,\beta_T$ represents the covariance schedule of the Gaussian noise in each step, and they are fixed constants.
    For the opposite direction of the Markov chain, we define a Markov chain with learned Gaussian transitions starting at $p(\boldsymbol x_T):= \mathcal{N}(\boldsymbol{x}_T; \boldsymbol{0},\boldsymbol{I})$. The disjoint distribution below is called the \emph{reverse process}.
    \begin{equation}
        p_{\theta}(\boldsymbol{x}_{0:T}):=p(\boldsymbol{x}_T) \prod_{t=1}^T p_{\theta}(\boldsymbol{x}_{t-1}| \boldsymbol{x}_t)
    \end{equation}
    in which 
    \begin{equation}
        p_\theta(\boldsymbol{x}_{t-1}|\boldsymbol{x}_t) := \mathcal{N}(\boldsymbol{x}_{t-1}; \boldsymbol{\mu}_{\theta}(\boldsymbol{x}_t,t), \boldsymbol{\Sigma}_\theta(\boldsymbol{x}_t,t))
    \end{equation}
    
    We want to minimize the negative log likelihood of the original graph $\boldsymbol x_0$. Using the similar variational bound with VAE, we have
    \begin{equation}
    \begin{aligned}
        \mathbb{E}[-\log p_{\theta} (\boldsymbol{x}_0)] &\le \mathbb{E}_q \left[-\log \frac{p_\theta (\boldsymbol{x}_{0:T})}{q(\boldsymbol{x}_{1:T}| \boldsymbol{x}_0)}\right] \\
        & = \mathbb{E}_q \left[ -\log p(\boldsymbol x _T) - \sum_{t\ge 1} \log\frac{p_\theta(\boldsymbol{x}_{t-1}|\boldsymbol{x}_t)}{q(\boldsymbol{x}_t|\boldsymbol{x}_{t-1})}\right]\\
        &=: \+L
    \end{aligned}
    \end{equation}
    
    Rewrite $\+L$ using KL divergence 
    \begin{equation}
    \begin{aligned}
        \+L &= D_{\text{KL}}(q(\boldsymbol x_T) \| p(\boldsymbol{x}_T))\\
        & + \mathbb{E}_{q}\left[\sum_{t\ge 1} D_{\text{KL}}(q(\boldsymbol x_{t-1} | \boldsymbol{x}_t)\| p_{\theta}(\boldsymbol{x}_{t-1} | \boldsymbol{x}_t) )\right]\\
        &+ H(\boldsymbol{x}_0)
    \end{aligned}
    \end{equation}
\subsubsection{ILVR}    
    Iterative Latent Variable Refinement \cite{ilvr} is a learning-free method of conditioning the generative process of the unconditional DDPM model to generate images that share high-level semantics from given reference images. With the condition $c$, we have
    \begin{equation}
        p_\theta(\boldsymbol x_{0:T}|c) = p(\boldsymbol x_T) \prod_{t=1}^T p_\theta (\boldsymbol x_{t-1} | \boldsymbol x_t , c)
    \end{equation}
    ILVR refine each unconditional transition $p_\theta(\boldsymbol{x}_{t-1}| \boldsymbol{x}_{t})$ with a downsample reference image. $\phi_N(\cdot)$ is a linear low-pass filtering operations, formed by a sequence of downsampling and upsampling by a factor of $N$. The condition $c$ is to ensure the downsampled image $\phi_N(x_0)$ of the generated image $x_0$ to be equal to $\phi_N(y)$ in which $y$ is the reference image. Then we have 
    \begin{equation}
        p_{\theta}(\boldsymbol{x}_{t-1}|\boldsymbol{x}_t,c) \approx p_\theta(\boldsymbol x_{t-1}| \boldsymbol{x}_t ,\phi_N(\boldsymbol{x}_{t-1}) = \phi_N(\boldsymbol y_{t-1}))
    \end{equation}
    The state $\boldsymbol{x}_t$ with condition $c$ is calculated as below:
    \begin{equation}
        \begin{aligned}
            &\boldsymbol x'_{t} \sim p_{\theta} (\boldsymbol{x}'_t | \boldsymbol{x}_{t+1}),\\
            & \boldsymbol x_{t} = \phi(\boldsymbol{y}_t) + (\boldsymbol{I} - \phi)(\boldsymbol{x}'_t)
        \end{aligned}
    \end{equation}
    $\boldsymbol{x'}_t$'s distribution is computed by the unconditional DDPM from state $\boldsymbol{x}_{t+1}$. By this way, we can calculate the generated image $\boldsymbol{x}_0$.
    
\begin{algorithm}
    \caption{ILVR for Face Swapping.}
    \begin{algorithmic}
        \Require Source image $x=I_S$, target image $y=I_T$ 
        \Ensure Result image $I_R$
        \State $\phi_N(\cdot)$: low-pass filter with scale N
        \State $\boldsymbol{z} \sim \+N (\boldsymbol{0},\boldsymbol{I})$
        \State $x_T \sim q\left(x_T \mid x\right)$
        \For {$t = T, \ldots, 1$}
            \State $\boldsymbol{z} \sim \+N (\boldsymbol{0},\boldsymbol{I})$
            \State $x_{t-1}' \sim p_{\theta}\left(x_{t-1}' \mid x_t\right)$
            \State $y_{t-1}  \sim q\left(y_{t-1} \mid y\right)$
            \State $x_{t-1} \gets \phi_N(y_{t-1}) +x_{t-1}' - \phi_{N}(x_{t-1}')$
        \EndFor
        \State $I_R \gets x_0$
        \State \Return $I_R$
    \end{algorithmic}
\end{algorithm}
    
    Since Gaussian noises still have much energy in lower frequencies, we had better minimize the difference between noises on $x_t$ and $y_t$. In the forward process, choose to use the same $\boldsymbol{z} \sim  \+N (\boldsymbol{0},\boldsymbol{I})$ to sample $x_{1..T}$ and $y_{1..T}$, which means that when we apply $\phi_N(\cdot)$ on both $\boldsymbol x_t$ and $\boldsymbol{y}_t$, the noises would have more similar outputs. That guarantees that the image signals passing through the filters will be restricted more upon the given conditions.

\section{Experiments} 
In this section, we use experiments to validate and compare the effectiveness of the invested methods. We evaluate the models on VGGFace2-HQ \cite{VGGFace2} dataset, which is a high resolution dataset of faces.

\paragraph{Implementation Details} The images are aligned and cropped into a standard position. We resize the images into $224\times 224$, except for the case of ILVR in which the pretrained DDPM only takes in $256\times 256$ images. As for the face recognition model in the ID Injection Module, we follow \cite{simswap} to use a pretrained Arcface \cite{arcface} model. For models with trainable face swapping procedure (\ie, CVAE with Identity Loss, CGAN \cite{simswap}, and CVAE-GAN), we train one intra-ID batch and then one inter-ID batch, alternately. Our models are trained for about 2 million iterations except for the DDPM, which pretrained on FFHQ \cite{ffhq} for 10 million iterations, and then finetuned on  VGGFace2-HQ \cite{VGGFace2} for another 1 million iterations. To select a representative model for each method, we benchmark all along the training process, and pick out the best model with lowest sum of ID retrieval and reconstruction loss. Due to the lack of time and capability to train a complete model thoroughly, we choose to finetune a pre-trained model of DDPM with ILVR method for another 1 milion iterations on VGGFace2-HQ \cite{VGGFace2} dataset. 

\subsection{Qualitative Results}
\begin{figure*}
  \centering
  \begin{subfigure}{0.45\linewidth}
    \includegraphics[width=\linewidth]{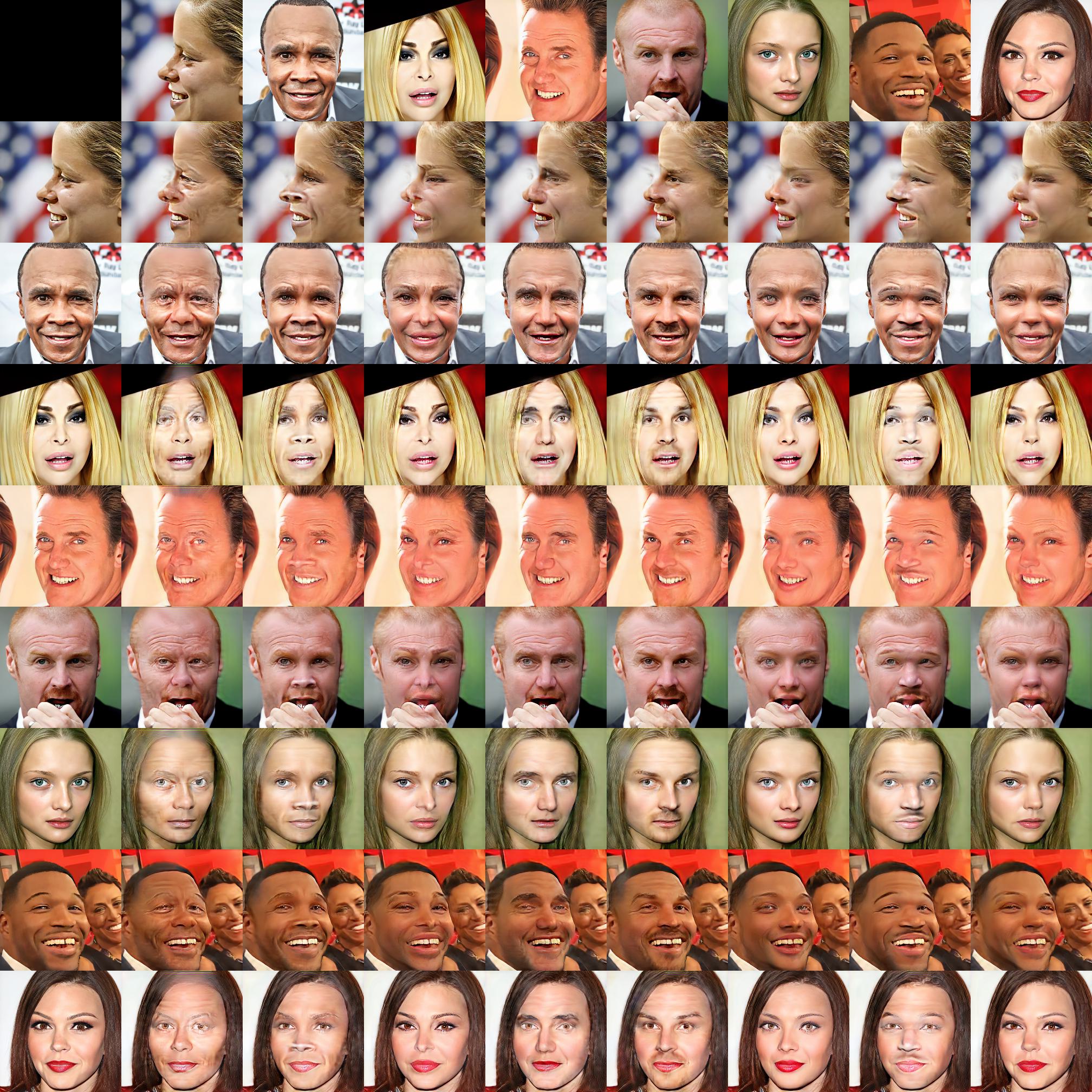}
    \caption{SimSwap}
    \label{fig:Qua-SimSwap}
    \end{subfigure}
    \hfill
    \begin{subfigure}{0.45\linewidth}
    \includegraphics[width=\linewidth]{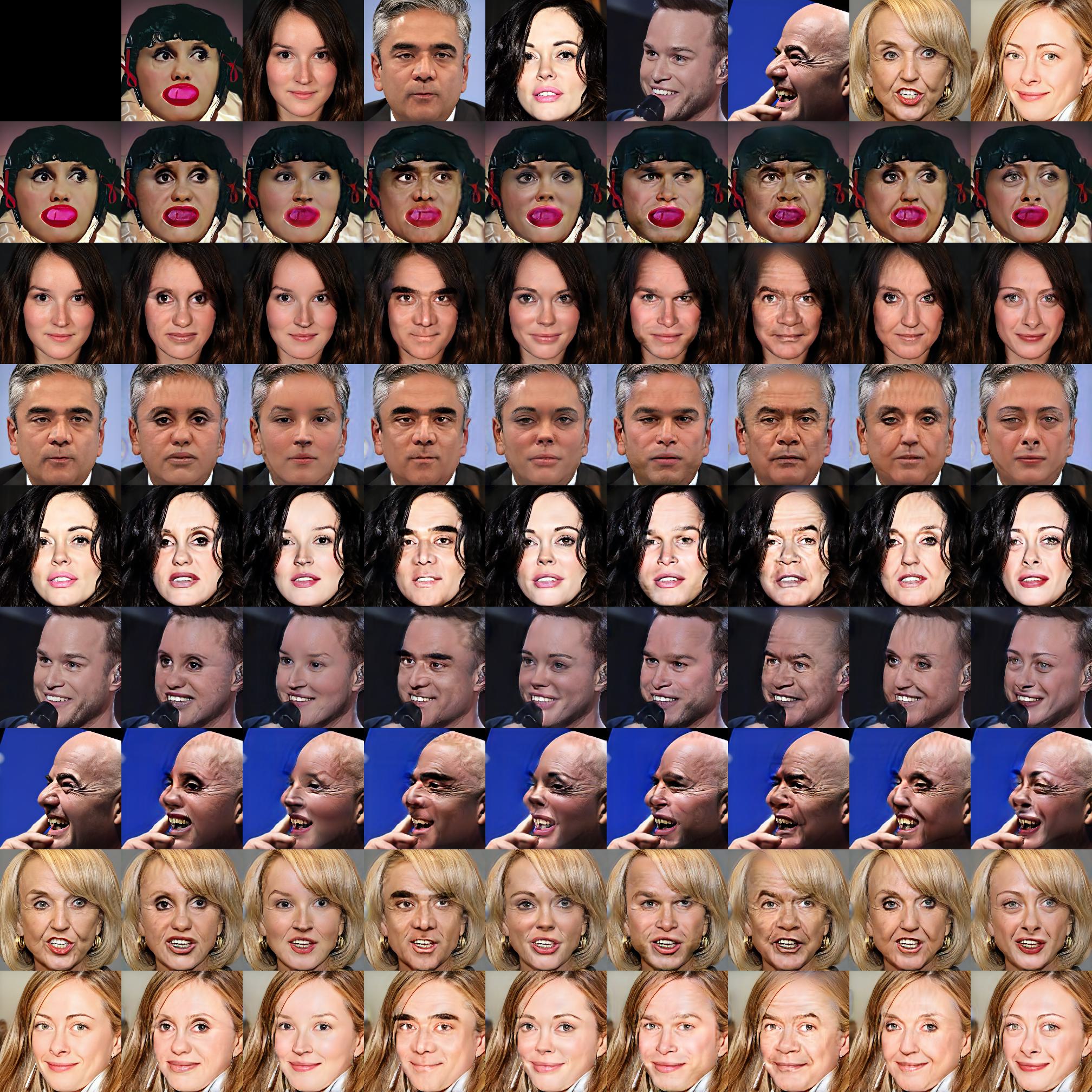}
    \caption{SimSwap-IG}
    \label{fig:Qua-SimSwap-IG}
    \end{subfigure}
    \hfill
    \begin{subfigure}{0.45\linewidth}
    \includegraphics[width=\linewidth]{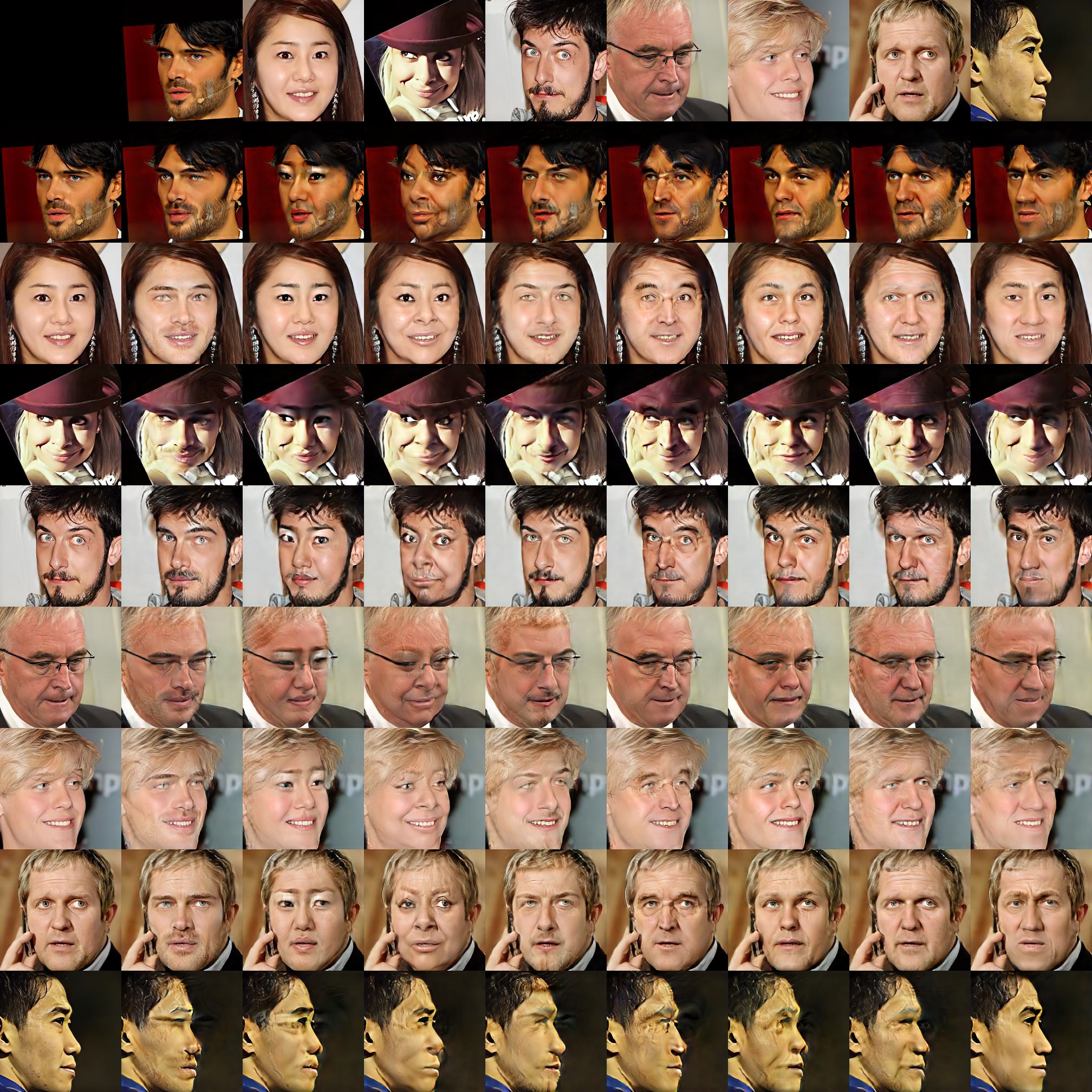}
    \caption{CVAE-GAN}
    \label{fig:Qua-CVAE-GAN}
    \end{subfigure}
    \hfill
    \begin{subfigure}{0.45\linewidth}
    \includegraphics[width=\linewidth]{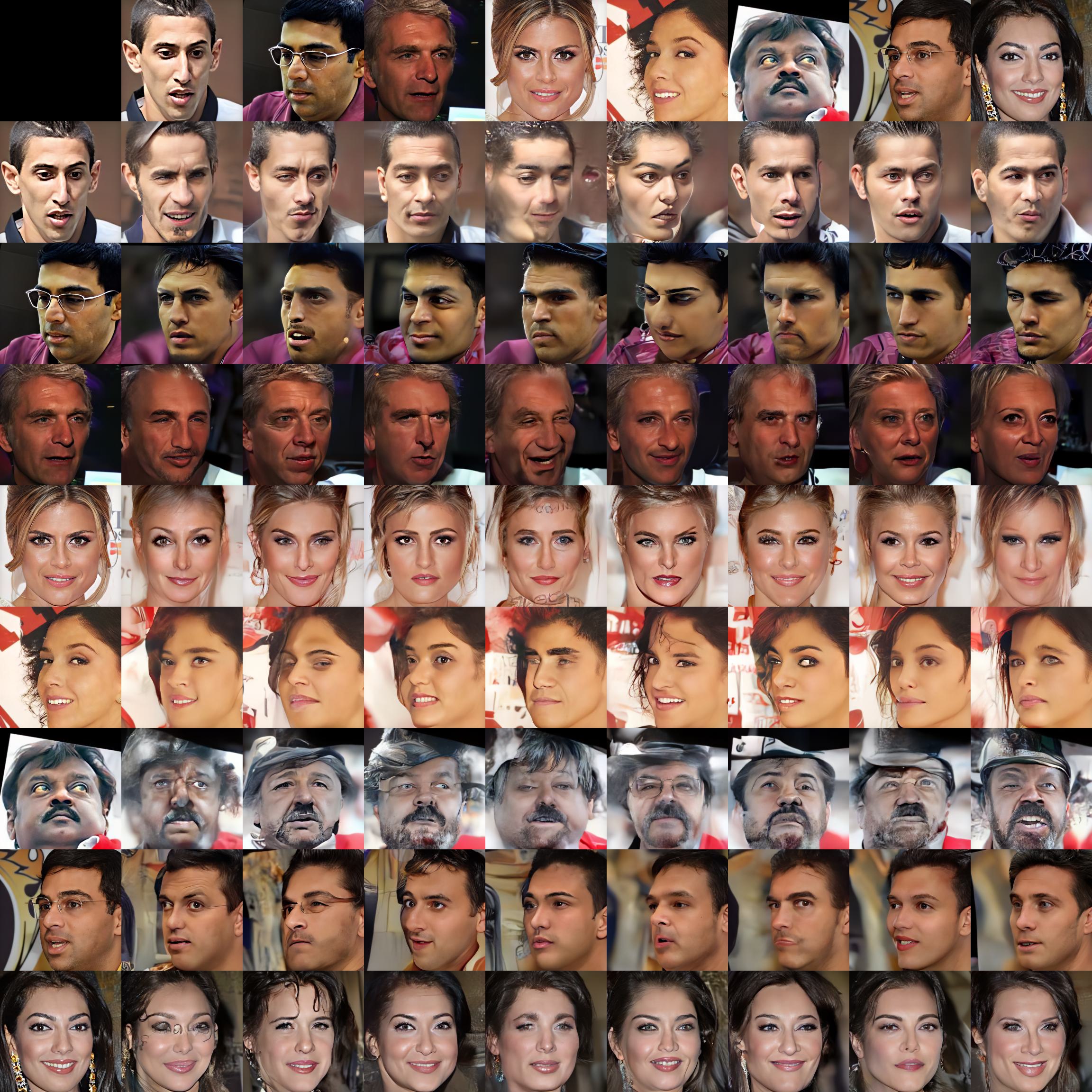}
    \caption{DDPM-ILVR}
    \label{fig:POS-DDPM-ILVR}
    \end{subfigure}
    \caption{Qualitative Results}
    \label{fig:Qualitative}
\end{figure*}
Let us show some qualitative face swapping results. As shown in \cref{fig:Qualitative}, here we pick 8 faces with different identities. We use them as both the source and the target image, and construct one face matrix a method. In each face matrix lays the face swapping results for all every source-target face pair.

Both GAN-based and DDPM-based methods are capable of preserving facial attributes (expression, gaze, posture and lighting). All our models manage to alter the identity of the picture significantly.

\subsection{Comparison between Methods}
We compare the models with by using them to swap the same image pairs respectively. Also, we rate these models quantitatively with identity loss and reconstruction loss. Furthermore, we follow \cite{simswap} introduce one more metric of identity retrieval. This metric is the $L2$ distance between normalized identity vector extracted by a face recognition model \cite{cosface} other than Arcface \cite{arcface}. This can avoid models trained with identity loss from cheating in evaluation on ability of identity injection. The evaluation results are recorded in \cref{tab:eval_compare}

\begin{table}
  \centering
  \begin{tabular}{@{}lccc@{}}
    \toprule
    Method &Recon Loss& ID Loss & ID Retrieval  \\
    \midrule
    SimSwap (baseline)& \textbf{0.006} & 0.008 & \textbf{0.012} \\
    CVAE & 0.007 & \textbf{0.005} & 0.020\\
    SimSwap-IG &   0.007 & 0.008 & 0.016\\
    CVAE-GAN-IG & \textbf{0.006} & 0.010 & 0.018 \\
    ILVR &0.043&0.031&0.029 \\
    \bottomrule
  \end{tabular}
  \caption{Benchmark on models with the best ID Retrieval for each method. IG stands for Identity Grouping. The best results are in \textbf{bold face}.}
  \label{tab:eval_compare}
\end{table}

For non-GAN CVAE models, we only benchmark the CVAE model with IDloss, which at least have the ability to swap faces.

\subsection{Extreme Examples}
Every method is not perfect and makes some trade-off. In other word, we can find inputs for which a certain method gives unsatisfying result. We call them extreme examples of a method. There exist both common extreme examples for all methods, and extreme examples for some specific methods.

\paragraph{Accessories}
Accessories like glasses, hats, and studs are removable and independent from personal identity. So they shall not get transferred to the target face in ideal face swapping. With ID extractor as the only identity input, which is pretrained to ignore accessories among images of an identical person, our CVAE and GAN model are free of such concern. However, ILVR takes in source face only, which means accessories from source face may remain in the result. Even worse, the diffusion model is prone to disturbance patterns. Accessories, especially glasses, sometimes appear from nothing. 
\subsection{Qualitative Results}
\begin{figure*}
  \centering
  \begin{subfigure}{0.45\linewidth}
    \includegraphics[width=\linewidth]{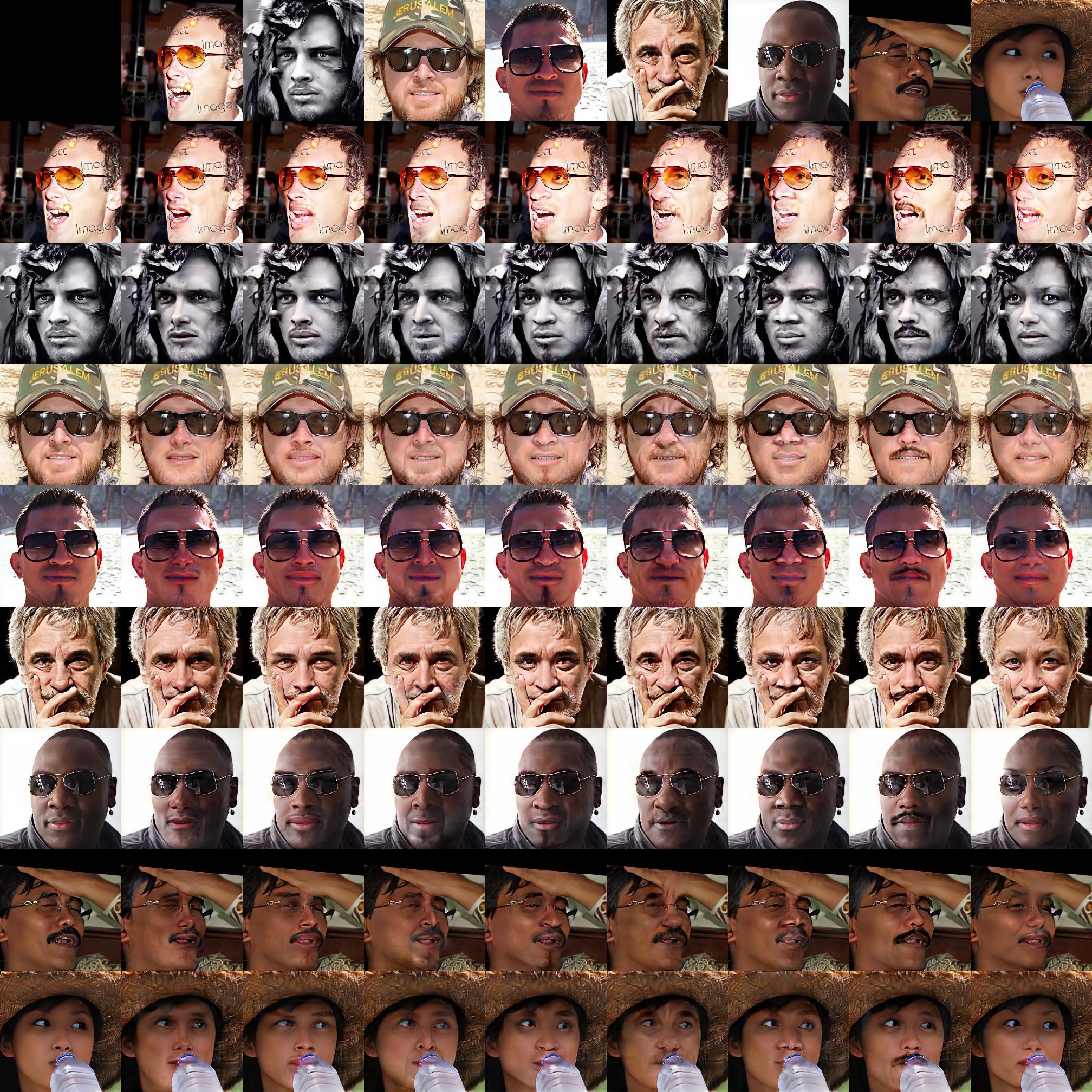}
    \caption{SimSwap}
    \label{fig:Acc-SimSwap}
    \end{subfigure}
    \hfill
    \begin{subfigure}{0.45\linewidth}
    \includegraphics[width=\linewidth]{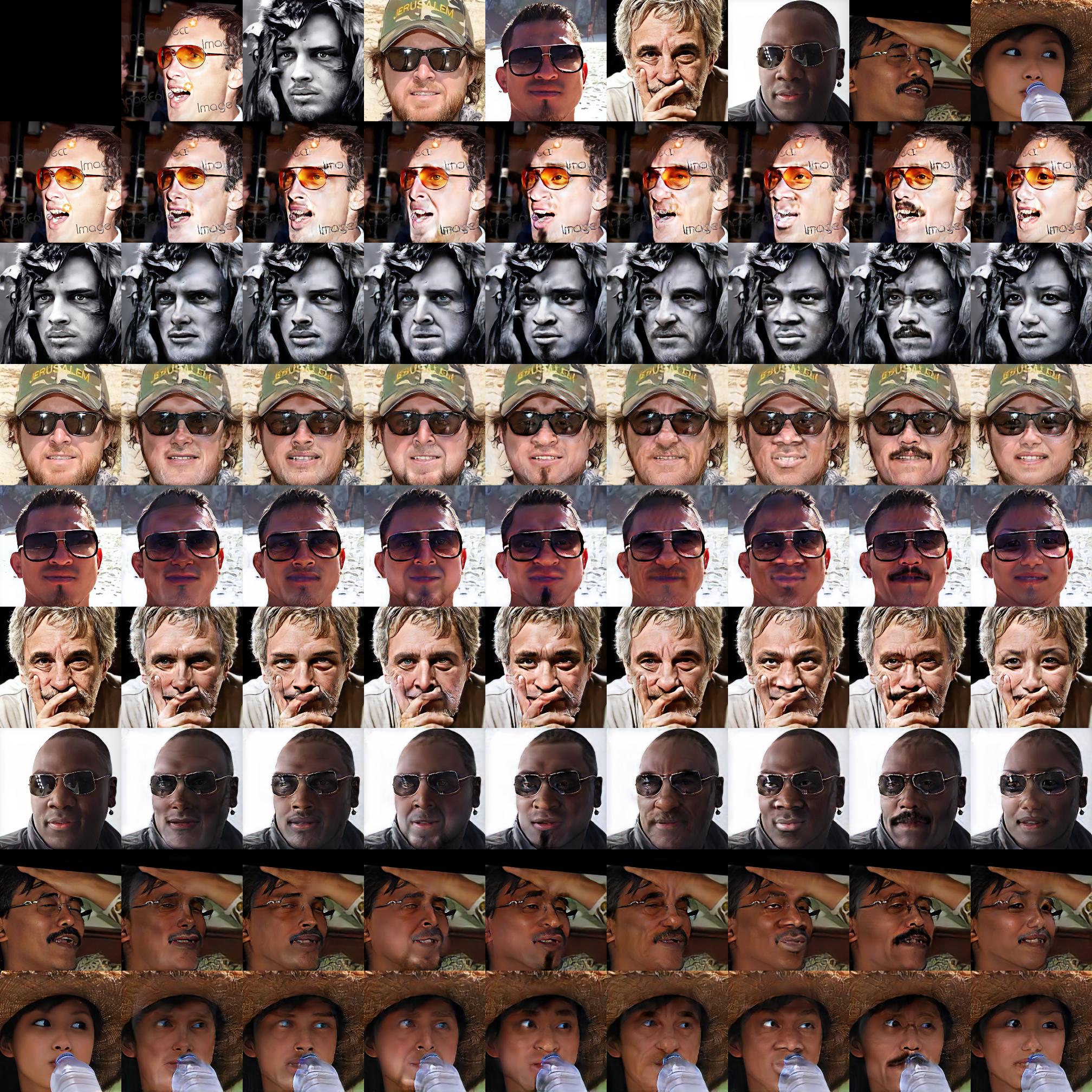}
    \caption{SimSwap-IG}
    \label{fig:Acc-SimSwap-IG}
    \end{subfigure}
    \hfill
    \begin{subfigure}{0.45\linewidth}
    \includegraphics[width=\linewidth]{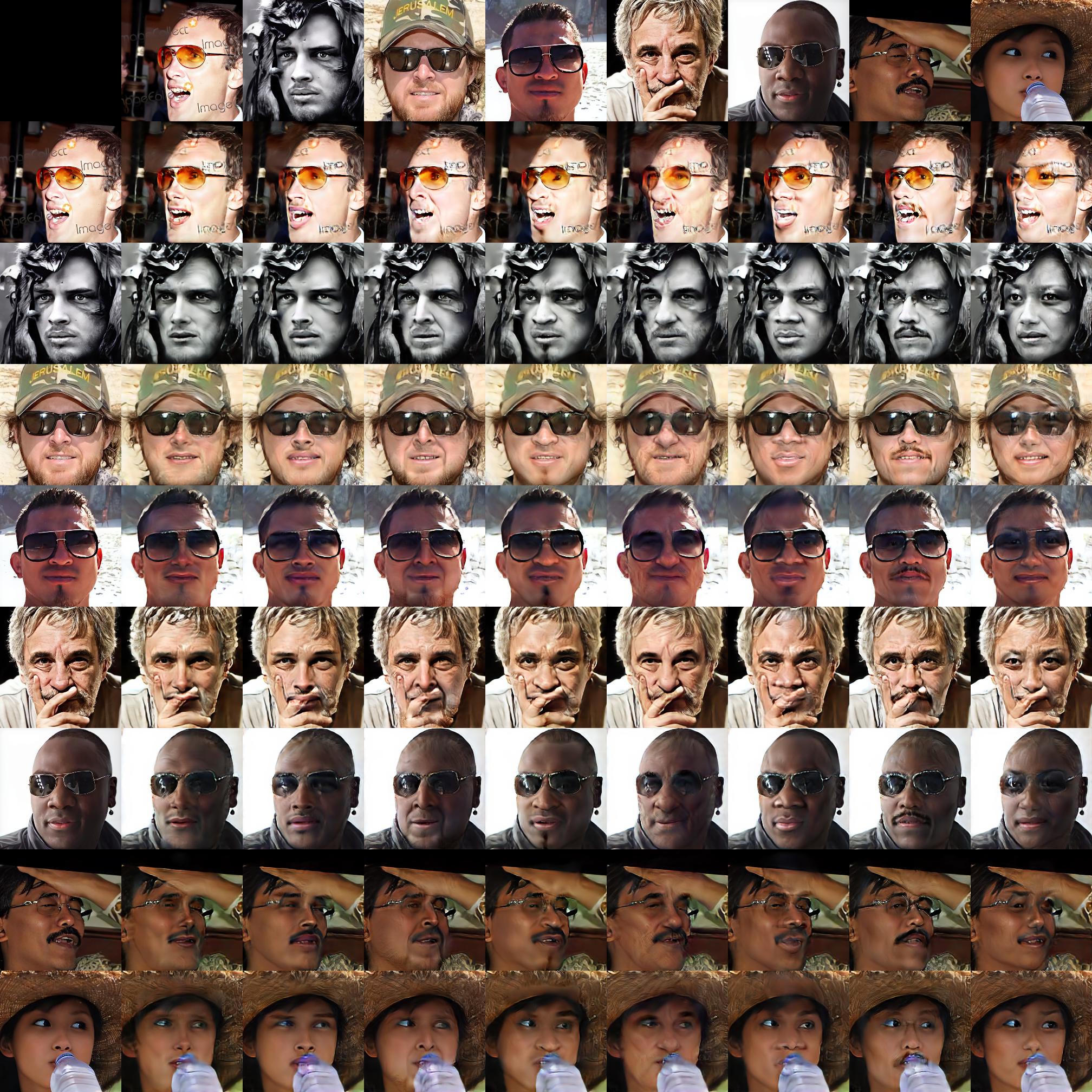}
    \caption{CVAE-GAN}
    \label{fig:Acc-CVAE-GAN}
    \end{subfigure}
    \hfill
    \begin{subfigure}{0.45\linewidth}
    \includegraphics[width=\linewidth]{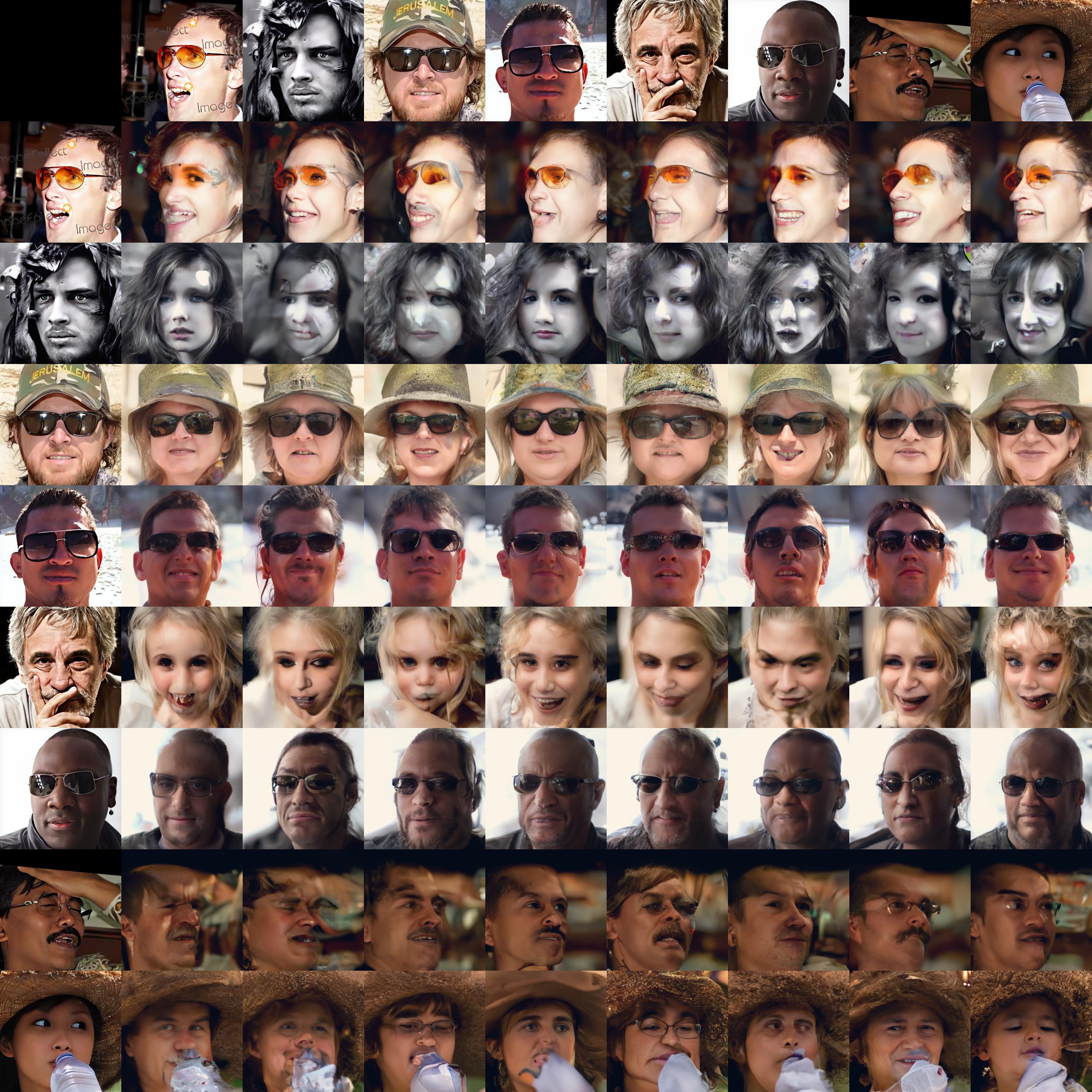}
    \caption{DDPM-ILVR}
    \label{fig:POS-DDPM-ILVR}
    \end{subfigure}
    \caption{Swapping Results of Faces with Accessories}
    \label{fig:Accesories}
\end{figure*}

\begin{figure*}
\centering
  \begin{subfigure}{0.45\linewidth}
    \includegraphics[width=\linewidth]{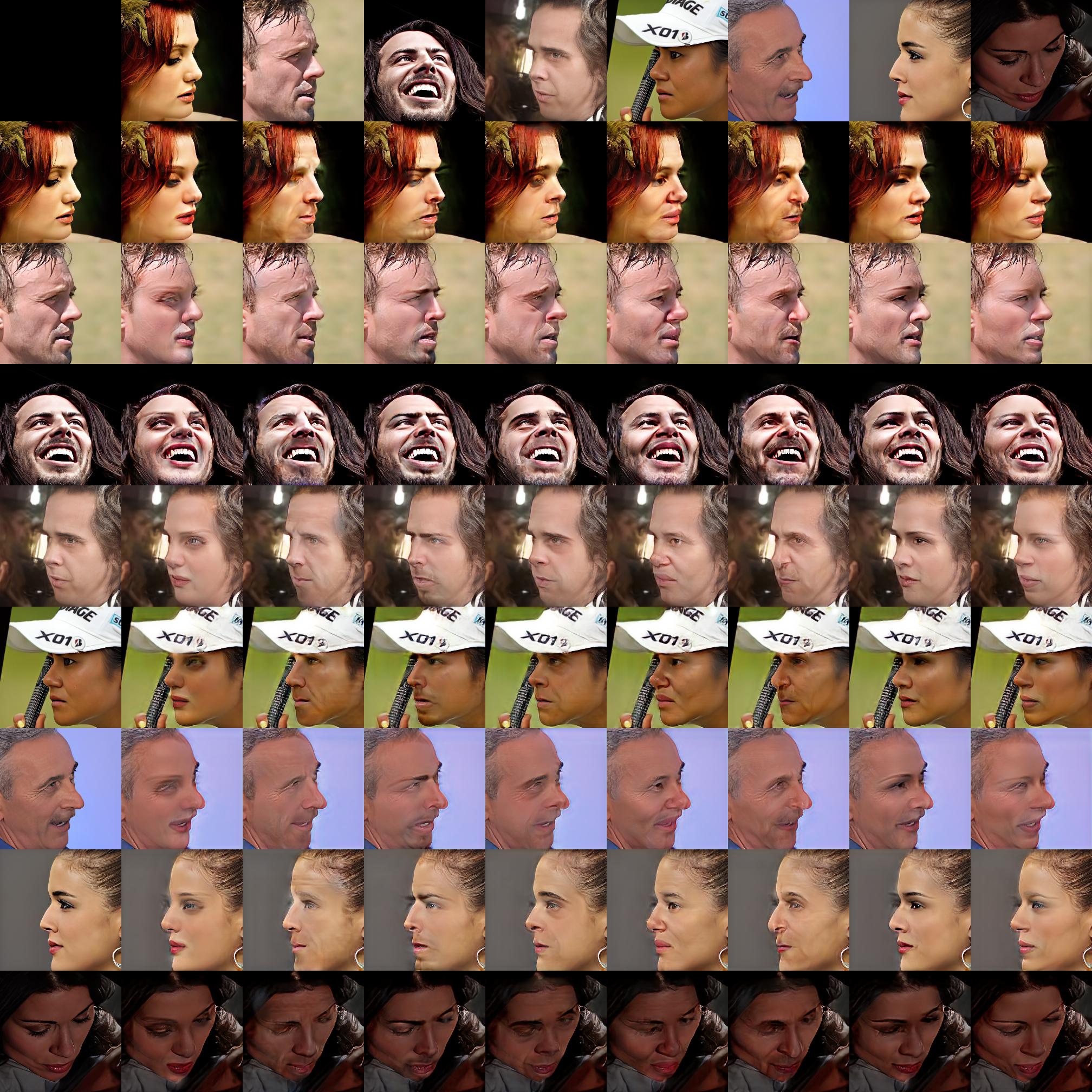}
    \caption{SimSwap}
    \label{fig:POS-SimSwap}
    \end{subfigure}
    \hfill
    \begin{subfigure}{0.45\linewidth}
    \includegraphics[width=\linewidth]{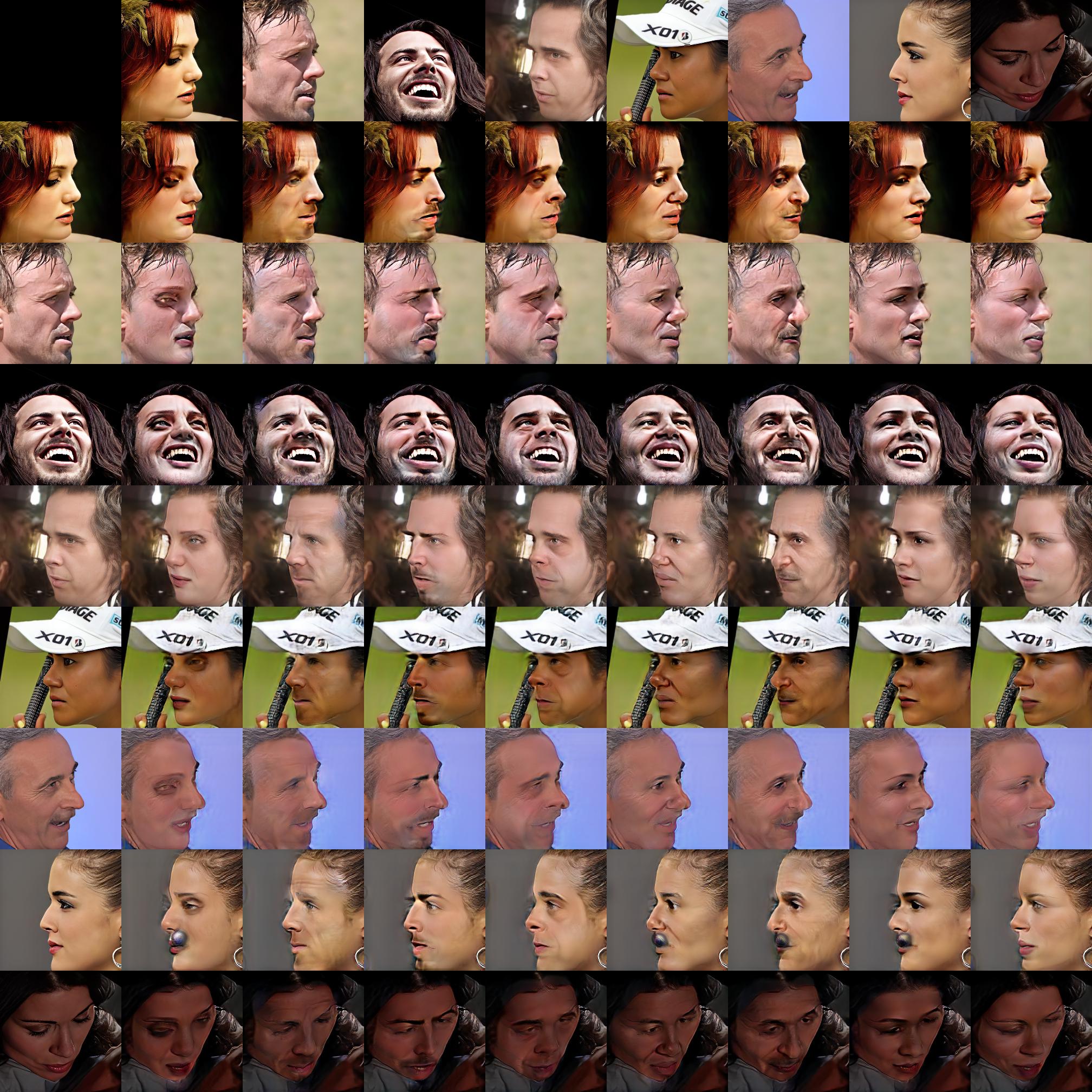}
    \caption{SimSwap-IG}
    \label{fig:POS-SimSwap-IG}
    \end{subfigure}
    \hfill
    \begin{subfigure}{0.45\linewidth}
    \includegraphics[width=\linewidth]{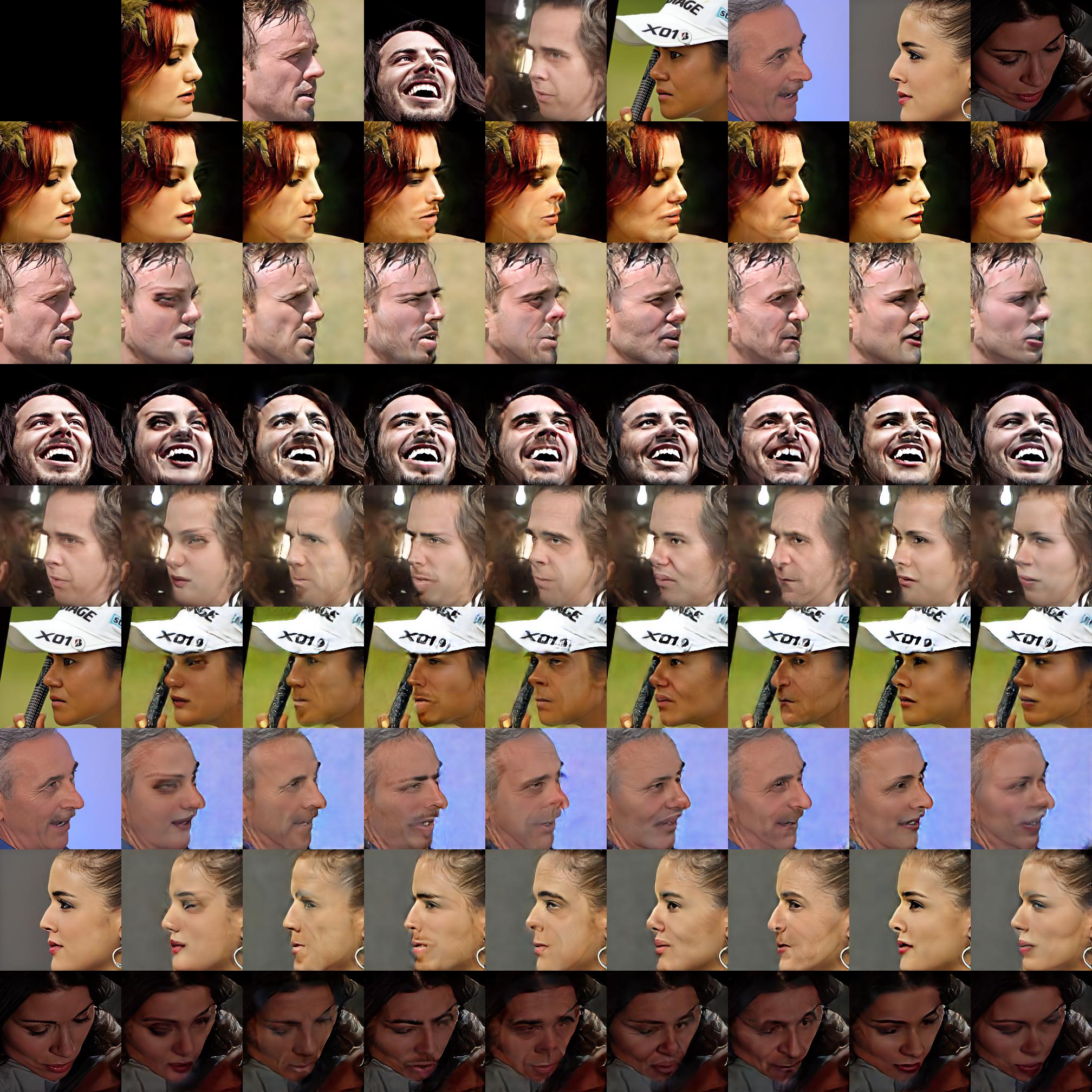}
    \caption{CVAE-GAN}
    \label{fig:POS-CVAE-GAN}
    \end{subfigure}
    \hfill
    \begin{subfigure}{0.45\linewidth}
    \includegraphics[width=\linewidth]{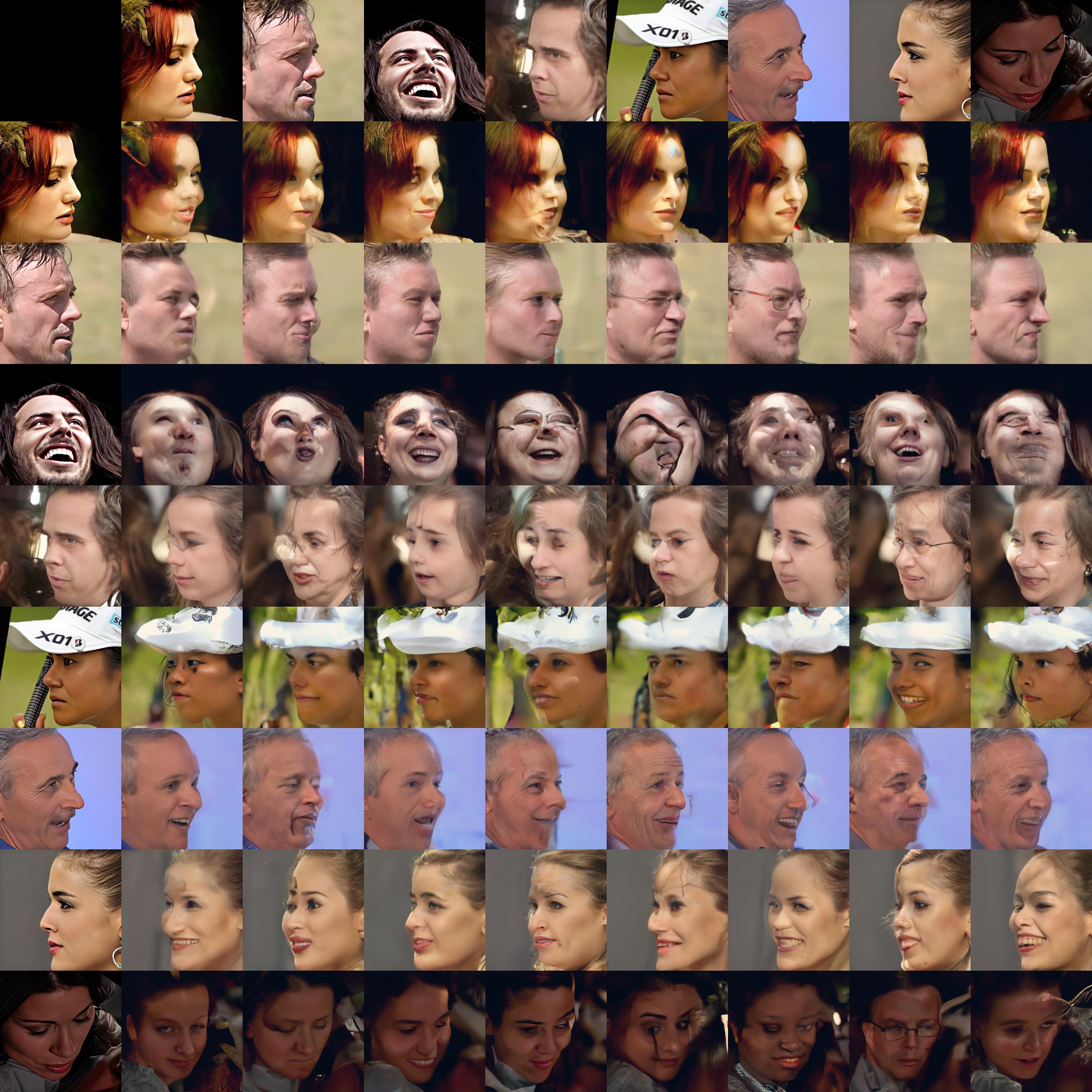}
    \caption{DDPM-ILVR}
    \label{fig:POS-DDPM-ILVR}
    \end{subfigure}
    \caption{Swapping Results of Faces with Extreme Postures}
    \label{fig:postures}
\end{figure*}

\paragraph{Postures}
Bad postures are fatal to non-GAN models. For the CVAE model, to minimize the ID loss, the model uses a kind of treating method that it "pastes" the identical objects of the source images onto the face of the targets with weird color. But if we removed the ID loss, the model will just aiming at reconstruction of the original image. The results of the models can be seen in \cref{fig:postures}

\begin{figure*}
\centering
  \begin{subfigure}{0.45\linewidth}
    \includegraphics[width=\linewidth]{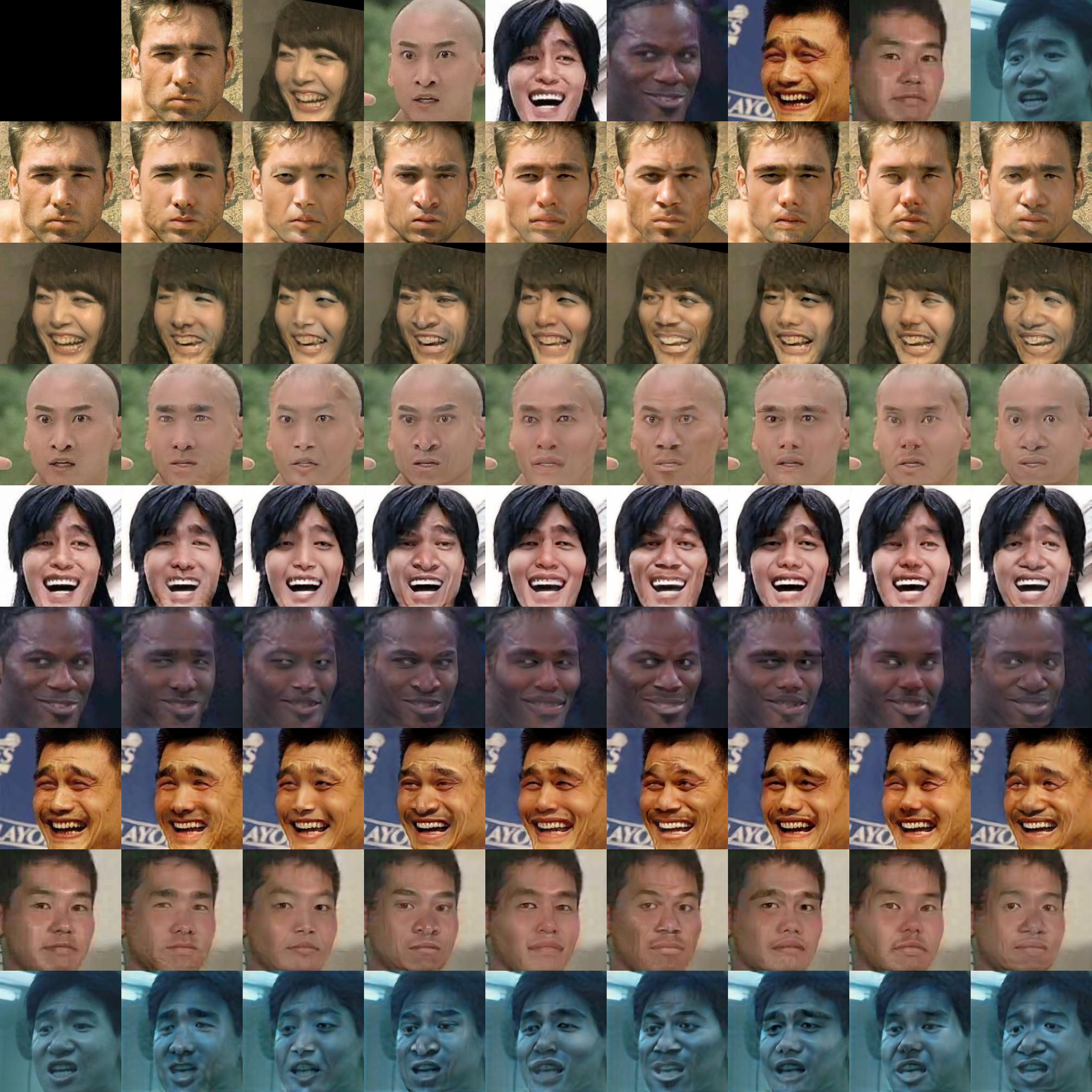}
    \caption{SimSwap}
    \label{fig:EXP-SimSwap}
    \end{subfigure}
    \hfill
    \begin{subfigure}{0.45\linewidth}
    \includegraphics[width=\linewidth]{figs/sample/Expressions/SimSwap_WO_intra-ID_random.jpg}
    \caption{SimSwap-IG}
    \label{fig:EXP-SimSwap-IG}
    \end{subfigure}
    \hfill
    \begin{subfigure}{0.45\linewidth}
    \includegraphics[width=\linewidth]{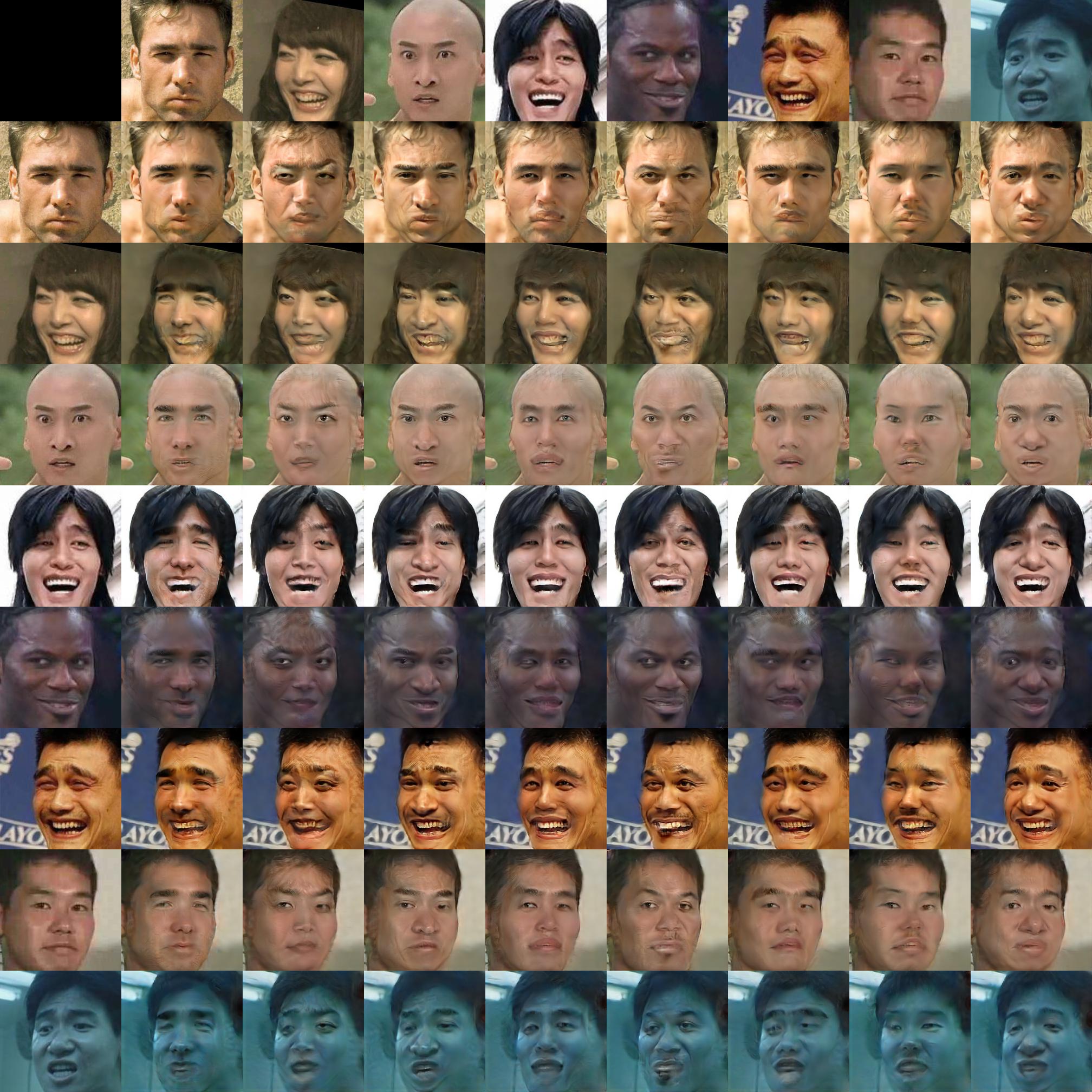}
    \caption{CVAE-GAN}
    \label{fig:EXP-CVAE-GAN}
    \end{subfigure}
    \hfill
    \begin{subfigure}{0.45\linewidth}
    \includegraphics[width=\linewidth]{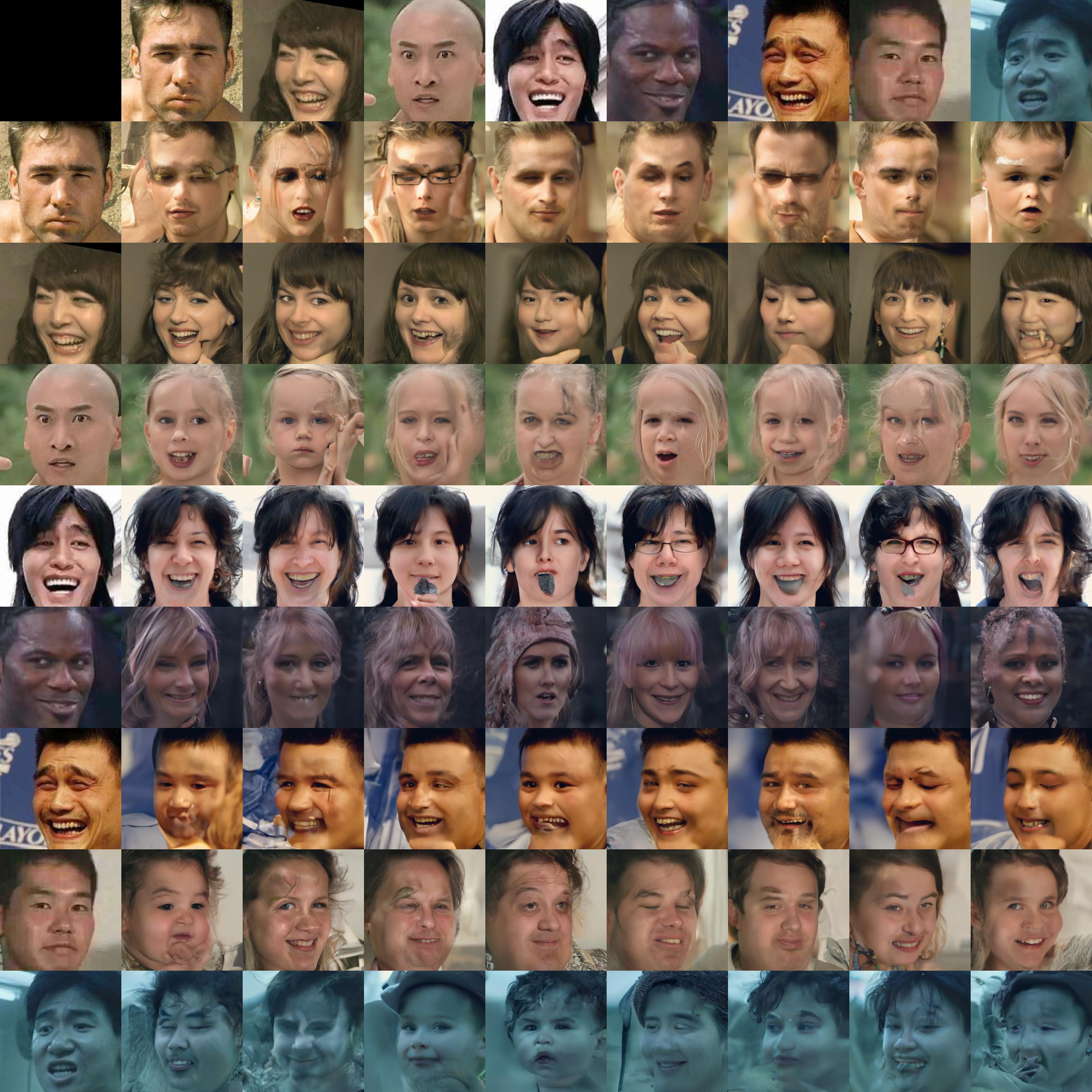}
    \caption{DDPM-ILVR}
    \label{fig:EXP-CVAE-GAN}
    \end{subfigure}
    \caption{Swapping Results of Faces with Extreme Expressions}
    \label{fig:expressions}
\end{figure*}

\paragraph{Expressions} 
Exaggerated expressions is tricky for identity extraction, attribute preservation, and even face detection. We refer to SimSwap for selection of such faces and experimented on our models. See \cref{fig:expressions} 

\section{Ablation Study and Analysis}
Some ablation experiments are done on the CVAE-GAN model. Furthermore, we also experimented the finetuning DDPM with different learning rate. All methods used are quite stable during training. No mode collapse did we see for any model.

\subsection{Ablation on SimSwap \cite{simswap}-like Models}
Since our experiments and models are heavily based on SimSwap \cite{simswap}, we put much attention on testing the performance of SimSwap.

\subsubsection{Ablation of FM Loss}
We follow SimSwap \cite{simswap} to make a comparision among SimSwap-oFM, SimSwap-$\overline{\text{wFM}}$, SimSwap-nFM  and SimSwap. SimSwap keeps the last few layers in the original Feature Matching term while removing the first few. SimSwap-oFM keeps all of the layers inthe original Feature Matching term. For SimSwap-$\overline{\text{wFM}}$, we keep the first few layers in the original Feature Matching term while removing the last few. And for SimSwap-nFM, we do not keep any layer in the Feature Matching term, which means the weight of FM loss is equal to $0$.

 As the data presented in \cref{tab:ablation}, we can find that the oFM and $\overline{\text{wFM}}$ performs better than the others. And considering the cost of calculation, the model with $\overline{\text{wFM}}$ loss is the best model according to this simple benchmark. If we remove the FM loss, the reconstruction loss becomes larger, which means that the generator may generate some weird images, and FM loss have the ability to restrain that phenomenon.
    
\begin{table*}
  \centering
  \begin{tabular}{@{}lccc|lccc@{}}
    \toprule
    Method &Recon Loss& ID Loss & ID Retrieval &  Method &Recon Loss& ID Loss & ID Retrieval\\
    \midrule
    SimSwap & 0.006 & 0.008 & \textbf{0.012} & SimSwap-IG & 0.007 & 0.008 & 0.016 \\
    SimSwap-oFM &\textbf{0.005} &0.008 &\textbf{0.012} & SimSwap-IG-oFM & 0.006 & 0.006 & 0.016\\
    SimSwap-nFM & 0.008 & \textbf{0.005} & 0.013 & SimSwap-IG-nFM & 0.009 & 0.006 & 0.016 \\
    SimSwap-$\overline{\text{wFM}}$ & 0.006 & 0.007 & \textbf{0.012} & SimSwap-IG-$\overline{\text{wFM}}$ & 0.006 & 0.006 & 0.016\\
    \bottomrule
  \end{tabular}
  \caption{Benchmark on SimSwap models with the best ID Retrieval for each kind of FM loss. The best results are in \textbf{bold face}.}
  \label{tab:ablation}
\end{table*}

\subsubsection{Ablation of Identity Grouping and CVAE-GAN}
    In the experiments, we have run all the ablation experiments in both cases of with and without Identity Grouping. Also, our CVAE-GAN is a generator-modified version of SimSwap. We can compare it directly with the SimSwap models.
    
    As is shown in \cref{tab:ablation} and \cref{tab:eval_compare}, our attempts do not improve the performance of SimSwap. Identity Grouping seems to significantly hint the model the distribution of ID vectors extracted by Arcface \cite{arcface}, which, unfortunately, leads to overfitting (with IG, ID Loss gets lower while ID Retrieval goes up). As a supervisor, it seems we had better tell the generator more about its evaluations rather than characteristics if we do not want it gets fooled. As for CVAE-GAN, the reparameterization may add to uncertainty of the outputs. To get a specific face, we need more certainty. The KL Loss which trades quality for prettier distribution may also add to the restrictions on generation.
    
\subsection{Ablation on ILVR \cite{ilvr}}
\begin{table}
  \centering
  \begin{tabular}{@{}lccc@{}}
    \toprule
    Method &Recon Loss& ID Loss & ID Retrieval \\
    \midrule
    Pretrained & 0.043 & 0.031 & 0.029 \\
    Finetuned (LR=1e-4) &0.043 &0.031 &0.029 \\
    Finetuned (LR=5e-5)& 0.043&0.031  &0.029  \\
    Finetuned (LR=1e-5) &0.043 & 0.031 & 0.029  \\
    Finetuned (LR=5e-6)& 0.043 & 0.031 & 0.029 \\
    \bottomrule
  \end{tabular}
  \caption{Benchmark on ILVR with different DDPM models.}
  \label{tab:ilvr}
\end{table}

From \cref{tab:ilvr}, it seems that the pretrained model is already powerful enough and has the ability to recover diffused images in the field of the dataset \cite{VGGFace2}. As is mentioned above, the model is trained on FFHQ \cite{ffhq} dataset, which means DDPM excels in generate images with high fidelity regardless of data field.
    
    
\section{Conclusion}
    To be honest, this work is just a superficial survey and some experiments on generative models. We have strictly followed the framework of SimSwap to build the network of CGAN. In the training phase, we introduced Identity Grouping to regularize the data loading, which, however leads to overfitting. From this we conclude that we shall not reveal to much about supervisors to the generator. Also, we get an extra part added to change the generator into a CVAE model and finally build a CVAE-GAN model. We finetune a pre-trained DDPM model on our dataset so as to make it suitable for our task. And finally we do some ablation experiments for our CVAE-GAN model. 
    
    \paragraph{Future Improvements}Based on what we have learned and explored, we come up with some na\"{i}ve ideas for future improvements. Some of them are the components we didn't apply, and some of them are the problem we met in the exploration. In this work, our preprocessing is just to calculate all of the ID vectors, may be there is a better preprocessing to fasten the process of train. For this task, we have the premise that faces are i.i.d.. For datasets with totally different distributions, methods based on DDPM seems promising to accomplish field transfer. Besides, we should take a deeper insight into the projected discriminator and the classifier that we didn't use in this work. Additionally, since DDPM is an excellent generator, perhaps we shall try DDPM-GAN in which DDPM and trainable face swapping pipeline work as the generator. For the CVAE-GAN model, we could try different weight for KL Loss. And for both Identity Grouping and CVAE-GAN, we could utilize this trick for criteria that are harder to fool. We have noticed that the ID Loss stops to drop early for SimSwap model, while it keeps going straight down and reaches a rather low value (though ID Retrieval increases). They could be some keys to breakthrough certain bottlenecks of the performance of a CGAN. By the way, the DDPM-based methods are extremely slow. In video faking or even real-time face swapping tasks, methods like conditioned noise level may help.
    








\section*{Acknowledgements}
The work was under the fancy instruction of Prof. Bingbing Ni and TA Yuhan Li. We also got help from Chen, who is one of the authors of SimSwap \cite{simswap}.
{\small
\bibliographystyle{ieee_fullname}
\bibliography{main}

\begin{thebibliography}{10}\itemsep=-1pt

\bibitem{wgan}
Martin Arjovsky, Soumith Chintala, and Léon Bottou.
\newblock Wasserstein gan, 2017.

\bibitem{cvaegan}
Jianmin Bao, Dong Chen, Fang Wen, Houqiang Li, and Gang Hua.
\newblock Cvae-gan: fine-grained image generation through asymmetric training.
\newblock In {\em Proceedings of the IEEE international conference on computer
  vision}, pages 2745--2754, 2017.

\bibitem{hingeloss}
Andrew Brock, Jeff Donahue, and Karen Simonyan.
\newblock Large scale gan training for high fidelity natural image synthesis.
\newblock {\em arXiv preprint arXiv:1809.11096}, 2018.

\bibitem{simswap}
Renwang Chen, Xuanhong Chen, Bingbing Ni, and Yanhao Ge.
\newblock Simswap: An efficient framework for high fidelity face swapping.
\newblock In {\em {MM} '20: The 28th {ACM} International Conference on
  Multimedia}, 2020.

\bibitem{ilvr}
Jooyoung Choi, Sungwon Kim, Yonghyun Jeong, Youngjune Gwon, and Sungroh Yoon.
\newblock Ilvr: Conditioning method for denoising diffusion probabilistic
  models.
\newblock {\em arXiv preprint arXiv:2108.02938}, 2021.

\bibitem{deepfakes}
Deepfakes.
\newblock Faceswap.
\newblock \url{https://github.com/deepfakes/faceswap}, 2017.

\bibitem{arcface}
Jiankang Deng, Jia Guo, Niannan Xue, and Stefanos Zafeiriou.
\newblock Arcface: Additive angular margin loss for deep face recognition.
\newblock In {\em Proceedings of the IEEE/CVF conference on computer vision and
  pattern recognition}, pages 4690--4699, 2019.

\bibitem{gan}
Ian Goodfellow, Jean Pouget-Abadie, Mehdi Mirza, Bing Xu, David Warde-Farley,
  Sherjil Ozair, Aaron Courville, and Yoshua Bengio.
\newblock Generative adversarial nets.
\newblock {\em Advances in neural information processing systems}, 27, 2014.

\bibitem{wgan-gp}
Ishaan Gulrajani, Faruk Ahmed, Martin Arjovsky, Vincent Dumoulin, and Aaron~C
  Courville.
\newblock Improved training of wasserstein gans.
\newblock In I. Guyon, U.~Von Luxburg, S. Bengio, H. Wallach, R. Fergus, S.
  Vishwanathan, and R. Garnett, editors, {\em Advances in Neural Information
  Processing Systems}, volume~30. Curran Associates, Inc., 2017.

\bibitem{resnet}
Kaiming He, Xiangyu Zhang, Shaoqing Ren, and Jian Sun.
\newblock Deep residual learning for image recognition.
\newblock In {\em Proceedings of the IEEE conference on computer vision and
  pattern recognition}, pages 770--778, 2016.

\bibitem{ddpm}
Jonathan Ho, Ajay Jain, and Pieter Abbeel.
\newblock Denoising diffusion probabilistic models.
\newblock {\em Advances in Neural Information Processing Systems},
  33:6840--6851, 2020.

\bibitem{adain}
Xun Huang and Serge Belongie.
\newblock Arbitrary style transfer in real-time with adaptive instance
  normalization.
\newblock In {\em Proceedings of the IEEE international conference on computer
  vision}, pages 1501--1510, 2017.

\bibitem{ffhq}
Tero Karras, Samuli Laine, and Timo Aila.
\newblock A style-based generator architecture for generative adversarial
  networks.
\newblock In {\em Proceedings of the IEEE/CVF conference on computer vision and
  pattern recognition}, pages 4401--4410, 2019.

\bibitem{vae}
Diederik~P Kingma and Max Welling.
\newblock Auto-encoding variational bayes.
\newblock {\em arXiv preprint arXiv:1312.6114}, 2013.

\bibitem{cgan}
Mehdi Mirza and Simon Osindero.
\newblock Conditional generative adversarial nets.
\newblock {\em arXiv preprint arXiv:1411.1784}, 2014.

\bibitem{cosface}
MuggleWang.
\newblock Cosface\_pytorch.
\newblock \url{https://github.com/MuggleWang/CosFace_pytorch}, 2018.

\bibitem{VGGFace2}
NNNNAI and neuralchen.
\newblock Vggface2 hq.
\newblock \url{https://github.com/NNNNAI/VGGFace2-HQ}, 2021.

\bibitem{ldm}
Robin Rombach, Andreas Blattmann, Dominik Lorenz, Patrick Esser, and Björn
  Ommer.
\newblock High-resolution image synthesis with latent diffusion models, 2021.

\bibitem{dpm}
Jascha Sohl-Dickstein, Eric Weiss, Niru Maheswaranathan, and Surya Ganguli.
\newblock Deep unsupervised learning using nonequilibrium thermodynamics.
\newblock In {\em International Conference on Machine Learning}, pages
  2256--2265. PMLR, 2015.

\bibitem{cvae}
Kihyuk Sohn, Honglak Lee, and Xinchen Yan.
\newblock Learning structured output representation using deep conditional
  generative models.
\newblock {\em Advances in neural information processing systems}, 28, 2015.

\bibitem{pix2pix}
Ting-Chun Wang, Ming-Yu Liu, Jun-Yan Zhu, Andrew Tao, Jan Kautz, and Bryan
  Catanzaro.
\newblock High-resolution image synthesis and semantic manipulation with
  conditional gans.
\newblock In {\em Proceedings of the IEEE conference on computer vision and
  pattern recognition}, pages 8798--8807, 2018.

\end{thebibliography}
}

\end{document}